\documentclass[10pt]{article}
\usepackage{natbib}
\usepackage{graphicx}
\usepackage{comment}
\usepackage{amsmath}
\usepackage{amssymb}
\usepackage{subfig}
\usepackage{booktabs}
\usepackage{enumerate}
\usepackage[ruled,vlined]{algorithm2e}
\usepackage{hyperref}
\usepackage{array}

\usepackage{amsmath,amsfonts,bm}

\def\eqref#1{equation~\ref{#1}}

\def\1{\bm{1}}

\newcommand{\test}{\mathcal{D_{\mathrm{test}}}}

\DeclareMathAlphabet{\mathsfit}{\encodingdefault}{\sfdefault}{m}{sl}
\SetMathAlphabet{\mathsfit}{bold}{\encodingdefault}{\sfdefault}{bx}{n}

\DeclareMathOperator*{\argmin}{arg\,min}

\usepackage[accsupp]{axessibility}  

\usepackage{enumitem}
\usepackage[preprint]{tmlr}
\author{\name Thibaut Issenhuth \email t.issenhuth@criteo.com \\
      \addr Criteo AI Lab, Paris, France \\
      LIGM, Ecole des Ponts, Univ Gustave Eiffel, CNRS, Marne-la-Vallée, France
      \AND
      \name Ugo Tanielian \email u.tanielian@criteo.com \\
      \addr Criteo AI Lab, Paris, France
      \AND \name Jérémie Mary
      \email j.mary@criteo.com \\
      \addr Criteo AI Lab, Paris, France
      \AND
      \name David Picard \email david.picard@enpc.fr\\
      \addr LIGM, Ecole des Ponts, Univ Gustave Eiffel, CNRS, Marne-la-Vallée, France}

\begin{document}
\newcolumntype{N}{>{\centering\arraybackslash}m{.5in}}

\title{EdiBERT: a generative model for image editing}
\maketitle

\begin{abstract}
       Advances in computer vision are pushing the limits of image manipulation, with generative models sampling highly-realistic detailed images on various tasks. However, a specialized model is often developed and trained for each specific task, even though many image edition tasks share similarities. In denoising, inpainting, or image compositing, one always aims at generating a realistic image from a low-quality one. In this paper, we aim at making a step towards a unified approach for image editing. 
       To do so, we propose EdiBERT, a bidirectional transformer that re-samples image patches conditionally to a given image. Using one generic objective, we show that the model resulting from a single training matches state-of-the-art GANs inversion on several tasks: image denoising, image completion, and image composition. We also provide several insights on the latent space of vector-quantized auto-encoders, such as locality and reconstruction capacities. The code is available at \href{https://github.com/EdiBERT4ImageManipulation/EdiBERT}{https://github.com/EdiBERT4ImageManipulation/EdiBERT}.
       
\end{abstract}

\section{Introduction}

\captionsetup[subfigure]{labelformat=empty}
\begin{figure}[ht]
\begin{tabular*}{340pt}{@{\extracolsep{\fill}}ccccc}
Denoising & Completion &  Compositing & Scribble-edit & Crossover \\
\subfloat{\includegraphics[width=0.19\linewidth]{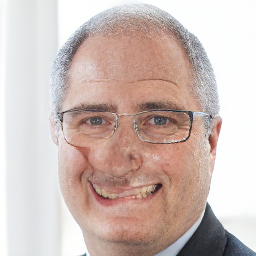}} &
\subfloat{\includegraphics[width=0.19\linewidth]{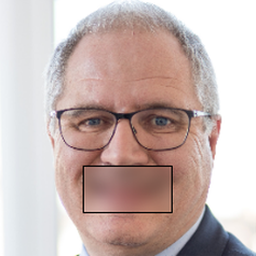}} &
\subfloat{\includegraphics[width=0.19\linewidth]{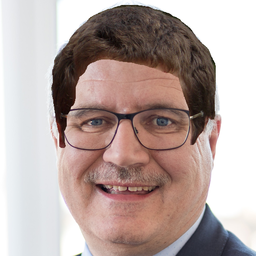}} &
\subfloat{\includegraphics[width=0.19\linewidth]{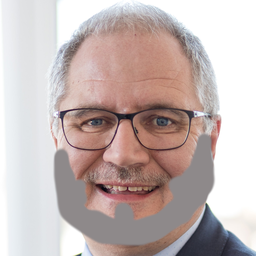}} &
\subfloat{\includegraphics[width=0.19\linewidth]{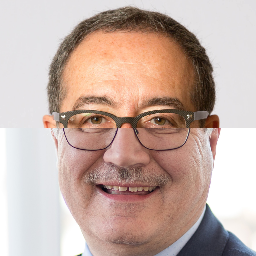}}\\
\subfloat{\includegraphics[width=0.19\linewidth]{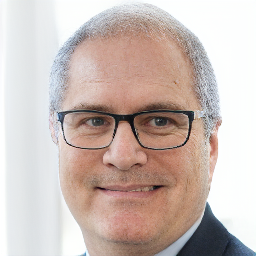}} &
\subfloat{\includegraphics[width=0.19\linewidth]{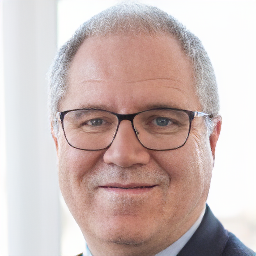}} &
\subfloat{\includegraphics[width=0.19\linewidth]{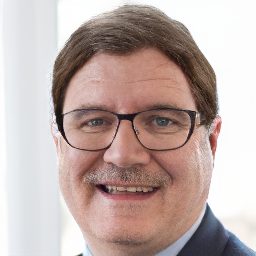}} &
\subfloat{\includegraphics[width=0.19\linewidth]{images/fig1_optim/fig1_scrib3.png}} &
\subfloat{\includegraphics[width=0.19\linewidth]{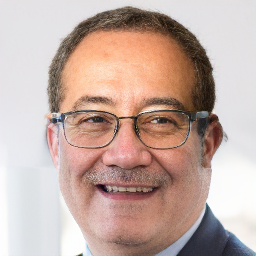}}\\
\subfloat{\includegraphics[width=0.19\linewidth]{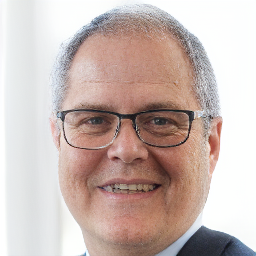}} &
\subfloat{\includegraphics[width=0.19\linewidth]{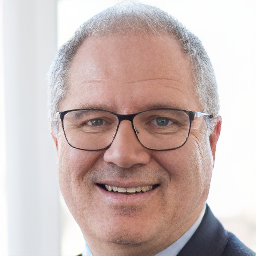}} &
\subfloat{\includegraphics[width=0.19\linewidth]{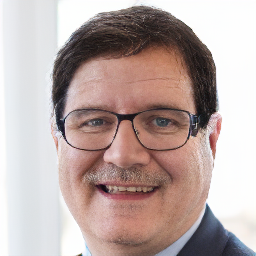}} &
\subfloat{\includegraphics[width=0.19\linewidth]{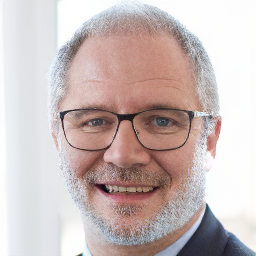}} &
\subfloat{\includegraphics[width=0.19\linewidth]{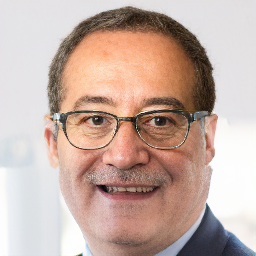}}\\
\end{tabular*}
\caption{Using a single and straightforward training, EdiBERT can tackle a wide variety of different tasks in image editing. The first row is the input, second and third rows are different samples from EdiBERT showing realism and consistency.   \label{fig:bert_editing}}
\end{figure} 

Significant progress in image generation has been made in the past few years, thanks notably to Generative Adversarial Networks (GANs) \citep{goodfellow2014generative}. In particular, the StyleGAN architecture \citep{karras2018style,karras2020analyzing} yields state-of-the-art results in data-driven unconditional generative image modeling. Empirical studies have shown the usefulness of such architecture when it comes to image manipulation. For example, by following specific directions in the latent space, one can modify an image attribute such as gender, age, the pose of a person \citep{shen2020interpreting}, or the angle \citep{jahanian2019steerability}. However, since the whole picture is generated from a Gaussian vector, changing some undesired elements while keeping the others frozen is difficult. To do so, edition algorithms involving optimization procedures have been proposed  \citep{abdal2019image2stylegan,abdal2020image2stylegan++} but with one main caveat: the results are not convincing when manipulating complex visuals \citep{niemeyer2021giraffe} (cf. experimental section for visual results). 

Independently, \citet{van2017neural} propose VQVAE, a promising latent representation by training an encoder/decoder  using  a discrete latent space. The authors demonstrate the possibility to  embed images in sequences of discrete tokens borrowing ideas from vector quantization (VQ), paving the way for the generation of images with autoregressive transformer models \citep{ramesh2021zero,esser2021taming}. On top of it, we argue that one of the benefits of this representation is that each token in the sequence is mostly coding for a localized patch of pixels (see section \ref{subsection:localVQ}), thus opening the possibility for an efficient localized latent edition. 

Aiming to build a unified approach for image manipulation, we propose a method that leverages both the spatial property of the discrete vector-quantized representation and the use of bidirectional attention models. In particular, we train a transformer based on ideas from the language model BERT \citep{devlin2018bert}, naming EdiBERT the resulting model. During training, EdiBERT tries to recover the original tokens of a perturbed sequence through a bidirectional attention schema. In computer vision, this approach has mainly been studied in the context of self-supervised learning \citep{bao2021beit, he2022masked}. We argue that training a single model from this generic objective provides a way to handle several editing tasks. To improve the visual quality and correct the information loss due to the quantization, we propose a post-processing procedure that better recovers the pixel content outside of the edited region.
With this unique training scheme and its companion sampling and post-processing algorithms, the obtained generative model can now be used in many different image manipulation tasks such as denoising, inpainting, or scribble-based editing, as shown in Figure \ref{fig:bert_editing}. 

To sum up, our contributions are the following:
\begin{itemize}
    \item[+] We analyze the VQ latent representations and illustrate their spatial properties, and show how to improve the reconstruction capabilities of VQGAN. This motivates the use of bidirectional transformers for image editing.
    \item[+] We show how to derive two different algorithms from a single model: one for image denoising where the locations of the edits are unknown, and a second one for inpainting or image compositing where a mask specifies the area to edit. \item[+] Finally, we show that using this generic and simple training algorithm and its companion post-processing allows us to achieve competitive results on various image manipulation tasks. 
\end{itemize}

\section{Related work}
We start this section by introducing transformer models for image generation before motivating the VQ representation and bidirectional models for image manipulation. 

\subsection{Autoregressive image generation}
The use of autoregressive transformer models \citep{vaswani2017attention} 
in computer vision has been made possible by two simultaneous research branches. First, the extensive deployment of attention mechanisms such as non-local means algorithms \citep{buades2005non}, non-local neural networks \citep{wang2018non}, and attention layers in GANs \citep{zhang2019self,pmlr-v139-hudson21a}. Second, the development of generative models sequentially inferring pixels via autoregressive convolutional networks such as PixelCNN \citep{van2016pixel,oord2016conditional}. Both of these works pave the way for the adoption of autoregressive transformers models in image generation \citep{parmar2018image}. The main interest of such formulation resides in a principled and tractable log-likelihood estimation of the empirical data. However, classical attention layers have a complexity scaling with the square of the sequence length, which is a bottleneck to scale to high-resolution images.

Recently, \citet{esser2021taming} were the first to apply autoregressive transformers on the discrete representation proposed by \citet{van2017neural}. In this framework, an encoder $E$, a decoder $D$, and a codebook/dictionary $Z$ are learned simultaneously to represent images with a single sequence of tokens. An autoregressive model is trained to generate these token sequences directly, stressing that high-capacity expressive transformers can generate realistic high-resolution images. To summarize, this framework consists of three main steps:
\begin{enumerate}
    \item Training a VQGAN, a set of encoder/decoder/codebook $(E, D, Z)$ with perceptual and adversarial losses.
    \item Training an autoregressive transformer to maximize the log-likelihood of the encoded sequences. 
    \item At inference, sampling a sequence with the transformer and decoding it with the decoder $D$.
\end{enumerate}
This representation also allowed to improve DALL-E \citep{ramesh2021zero}, a state-of-the-art text-to-image model translation.

\subsection{Bidirectional attention}
The main property of autoregressive models is that they only perform attention on previous tokens, making them inadequate when dealing with image manipulation \citep{esser2021imagebart}. Some works alleviate this bias in different ways. \citet{yang2019xlnet} learn an autoregressive model on random permutations of the ordering. \citet{cao2021image} propose a model where missing tokens are inferred autoregressively, conditionally to the set of kept tokens. Similarly, \citet{wan2021high} use an auto-regressive procedure conditioned on the masked image, while \citet{yu2021diverse} use BERT training with [MASK] tokens and Gibbs sampling. If this setting is ideal for tasks with masked tokens such as inpainting, it makes it ill-posed for scribble-editing and insertion without existing paired datasets. On the opposite, our EdiBERT tackles all tasks without any need for supervision. Finally, \citet{esser2021imagebart} train ImageBART, a transformer that reverts a diffusion process in the discrete latent space of VQGAN. Each generated sequence is conditioned on the previous one, thus paying attention to the whole image. However, this method is computationally heavy since it requires making $N\times L$ inferences, where $N$ is the number of generated sequences and $L$ the number of tokens in the sequence. Moreover, the authors do not the limited reconstruction capabilities of VQGANs. 
In this paper, we argue that by performing bidirectional attention over all the tokens (perturbed or not), it is now possible to train a single model tackling many editing tasks.


\subsection{Unifying image manipulation}
Initially, image manipulation methods were implemented without any trainable parameters. Image completion was tackled using nearest-neighbor techniques along with a large dataset of scenes \citep{hays2007scene}. As to image insertion, blending methods were widely used, such as the Laplacian pyramids \citep{burt1987laplacian}. In recent years, image manipulation has benefited from the advances of deep generative models. A first line of works has consisted of gathering datasets of corrupted and target images to train conditional generative models. By doing so, one can therefore learn a mapping from any corrupted image to a real one. This technique can be done using either an unpaired dataset \citep{zhu2017unpaired}, or a paired dataset \citep{zhao2020large}. For example, \citet{liu2021deflocnet} proposes an encoder-decoder architecture for sketch-guided image inpainting.  However, in all cases, a dataset with both types of images is required, therefore limiting the applicability. 

To avoid this dependency, a second idea  - known as GAN inversion methods - leverages pre-trained unconditional GANs. They work by projecting edited images on the manifold of real images learned by the pre-trained GAN. It can be solved either by optimization \citep{abdal2019image2stylegan,abdal2020image2stylegan++,pmlr-v139-daras21a}, or with an encoder mapping from the image space to the latent space \citep{chai2021using,richardson2021encoding,tov2021designing}. Pros of these GAN-based methods are that one benefits from outstanding properties of StyleGan, state-of-the-art in image generation. However, the main drawback of these methods is that they rely on a task-specific loss function that needs to be defined and optimized. Finally, and closer to our work is the SDEdit model proposed by \citet{meng2021sdedit} where the authors use Langevin's dynamics for image edition. However, the inference time of such model is still particularly slow. 

\section{Motivating EdiBERT for image editing}
This section  gives a global description of the proposed EdiBERT model. We start with notations before describing the different steps leading to the BERT-based edition. 

\subsection{Notations}
Let $I$ be an image with a width $w$, a height $h$ and a number $c$ of channels. $I$ thus belongs to $\mathbb{R}^{h\times w\times c}$. Also, let $(E, D, Z) $ be respectively the encoder, decoder and codebook defined in  VQVAE and VQGAN \citep{van2017neural,esser2021taming}. The codebook consists of a finite number of tokens with fixed vectors in an embedding space. Let $Z = \{t_1, \hdots, t_N \}$ with $t_k \in \mathbb{R}^d$ and $N$ being the size of the codebook. 

For any given image $I$, the encoder $E$ outputs a vector $E(I) \in \mathbb{R}^{l \times  d}$, which is then quantized and reshaped into a sequence $s$ of length $l$ as follows:
\begin{align}
    s = (\underset{z \in Z}{\text{arg min }} \| E(I)_1 - z \|, \hdots, \underset{z \in Z}{\text{arg min }} \| E(I)_l - z \|)) = Q_Z(E(I)),
\end{align} 
where $Q_Z$ refers to the quantization operation using the codebook $Z$. Recall that, after the quantization step, one gets a sequence $s \in Z^{l}$.

Let note $\mathcal{D}$, the available image dataset. From a pre-trained encoder $E$ and codebook $Z$, one can transform the image dataset $\mathcal{D}$ into a dataset of token-sequences $\mathcal{D}_S$:
\begin{equation}
    \mathcal{D}_S = \{ Q_Z(E(I)), I \in \mathcal{D} \}.
\end{equation}
Interestingly, when learning transformers on sequences of tokens, the practitioner is directly working with $\mathcal{D}_S$ and not $\mathcal{D}$.

For the rest of the paper, we consider $p_\theta$ a transformer model parameterized with $\Theta$. The present section aims to illustrate the chosen objective for the EdiBERT model on the dataset $\mathcal{D}_S$. For each position $i$ in $s$, we note $p_\theta^i(.|s)$, the modeled distribution of tokens conditionally to $s$.

\subsection{Learning sequences with autoregressive models}
Previous works training transformers model on $\mathcal{D}_S$ chose autoregressive models \citep{esser2021taming}. In this setting, for a given sequence $s=(s_1, \hdots, s_l) \in \mathcal{D}_S$ of discrete tokens, the likelihood $p_\theta(s)$ of the sequence $s$ is given using the following formula:
\begin{equation}
    p_\theta(s) = \prod_{i=1}^l \ p_\theta^i(s_i|s_{<i}), \quad \text{with } s_{<i} = (s_1, \hdots, s_{i-1}).    
\end{equation}
Since each $s_i \in Z$ is discrete, this decomposition is implemented thanks to a causal left-to-right attention mask and a softmax output layer. Finally, given a specified set of parameters $\Theta$, the objective of the autoregressive model is:
\begin{equation}\label{eq:autoregressive}
    \underset{\theta \in \Theta}{\text{arg max }}  \mathbb{E}_{s \in \mathcal{D}_s} \ \log p_\theta(s).
\end{equation}

\paragraph{Limitations of the model.}
If this setting is well suited for unconditional image generation, it is ill-posed for image manipulation tasks \citep{esser2021imagebart}. Indeed, in the case of scribble-based editing, one wants to resample tokens conditionally to the whole image. In the case of inpainting, one wants to resample tokens conditionally to a random subset of the tokens.  

\subsection{A unique training objective for EdiBERT.}
Let us define the training objective for EdiBERT. For any sequence $s = (s_1,...,s_L)$, a function $\varphi$ randomly selects a subset of $k$ indices $\{\varphi_1,...,\varphi_k\}$ where $\varphi_k< L$. At each selected position, a perturbation is applied on the token $s_{\varphi_i}$. We attribute a random token with probability $p$, or keep the same token with probability $1-p$. Consequently, the perturbed token $\Tilde{s}_{\varphi_i}$ becomes:
\begin{align}
    \Tilde{s}_{\varphi_i} &= \mathbb{U}(Z) \quad \text{with probability } p, \\
    \Tilde{s}_{\varphi_i} &= s_{\varphi_i} \quad \text{with probability } 1-p,
\end{align}
where $\mathbb{U}(Z)$ refers to the uniform distribution on the space of tokens $Z$.  Similarly to \citet{bao2021beit}, the sampling function $\varphi$ is defined with a 2D masking strategy, and the training positions are selected by drawing random 2D rectangles in the 2D patch extracted from the encoder. Note that, contrarily to \citet{bao2021beit} and \citet{devlin2018bert}, we only use random tokens from the codebook but no [MASK] tokens. We argue this setting corresponds more to our use cases (denoising or editing), where we want to sample conditionally to an entire perturbed sequence. 

Let us now call $\Tilde{s}$ and $\Tilde{\mathcal{D}_{s}}$, respectively the perturbed sequence and dataset:
\begin{align}
    \Tilde{s} = (s_1, \hdots, \Tilde{s}_{\varphi_1}, \hdots, \Tilde{s}_{\varphi_k}, \hdots, s_L), \quad \text{and} \quad 
    \Tilde{\mathcal{D}} = \{\Tilde{s}, \ s \in \mathcal{D} \}.
\end{align}
The training consists in the following objective :
\begin{equation}\label{eq:bidirectionnal}
    \underset{\theta \in \Theta}{\text{arg max }} \ \mathbb{E}_{\Tilde{s} \in \Tilde{\mathcal{D}_s}} \ \frac{1}{k} \ \sum_{i=1}^k \log p_\theta^i(s_{\varphi_i} | \Tilde{s}) .
\end{equation}
Contrary to \eqref{eq:autoregressive}, we note that the objective in \eqref{eq:bidirectionnal} enforces the model to perform attention over the whole input image. 

\paragraph{Sampling from EdiBERT:} \citet{wang2019bert} show that it is possible to generate realistic samples with a BERT model starting with a random initialization. However, compared with standard autoregressive language models, the authors stress that BERT generations are more diverse but of slightly worse quality. Building on the findings of \citet{wang2019bert}, we do not aim to use BERT for pure unconditional sequence generation but rather improve an already existing sequence of tokens. In our defined EdiBERT model, for any given position $i \in s$, a token will be sampled according to the multinomial distribution $p_\theta^i(.|s)$.

\subsection{On the locality of Vector Quantization encoding \label{subsection:localVQ}}
In this paper, we argue that one of the main advantages of EdiBERT comes from the VQ latent space proposed by \citet{van2017neural} where each image is encoded in a discrete sequence of tokens. In this section, we illustrate with simple visualizations the property of this VQGAN encoding. We explore the spatial correspondence between the position of the token in the sequence and a set of pixels for the encoded image. We aim at answering the following question: do local modifications of the image lead to local modifications of the latent representation and \textit{vice versa}?

\begin{figure}
\begin{center}
    \textbf{Modifying the latent space via the image space} \\
     {\includegraphics[width=0.45\linewidth]{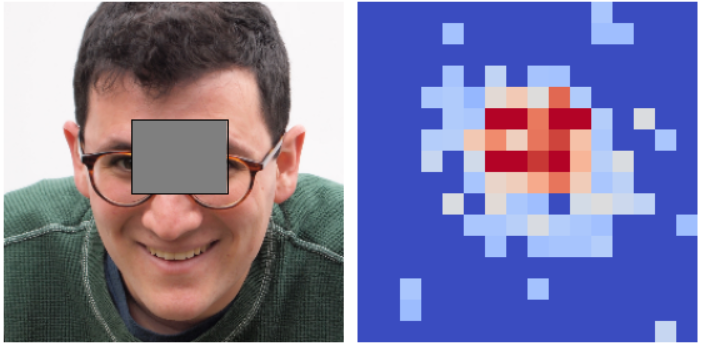}}
    \subfloat{\includegraphics[width=0.45\linewidth]{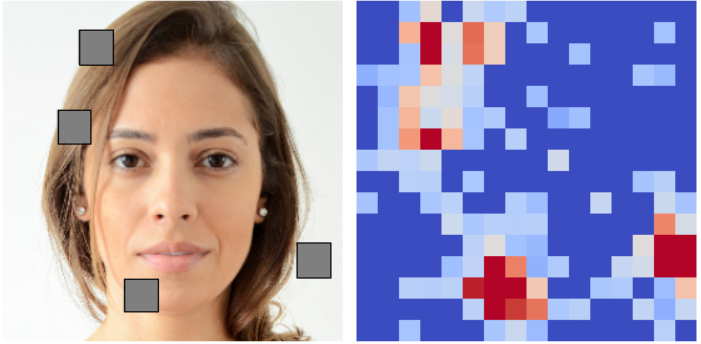}}\\
    \textbf{Modifying the image via the latent space} \\
    {\includegraphics[width=0.90\linewidth]{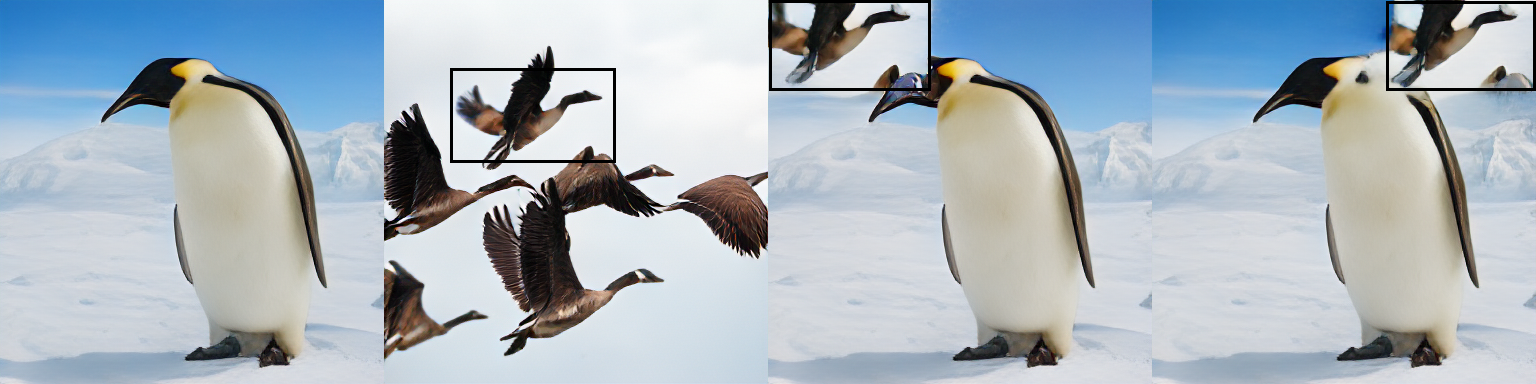}}\\
    \caption{Each VQGAN token is strongly tied to a small spatial area in the image space. First row: perturbed images and the respective variations of tokens in the latent space. Second row: collages of images can easily be done with collages of latent representations. \label{fig:vqgan_encoding}}
\end{center}
\end{figure}

\paragraph{Modifying the image.} To answer this question, images are voluntarily perturbed with grey masks ($i \longrightarrow i_m$). Then, we encode the two images, quantize their representation using a pre-trained codebook, and plot the distance between the two latent representations: $\|Q_Z(E(i)) - Q_Z(E(i_m)) \|_2^2 $. The results are shown in the first row in Figure \ref{fig:vqgan_encoding}. Due to the large receptive field of the encoder, tokens can be influenced by distant parts of the image: the down-sampled mask does not recover all of the modified tokens. However, tokens that are largely modified are either inside, or very close to the down-sampled mask. 

\paragraph{Modifying the latent space.} To understand the interchange of the correspondence between tokens and pixels, we want to stress how one can easily manipulate images via the discrete latent space. We show in Figure \ref{fig:vqgan_encoding} that extracting a specific area of a given source image and inserting it at a chosen location in another image is possible by replacing the corresponding tokens in the target sequence with the ones of the source. Consequently, this spatial correspondence between VQGANs latent space and image space is interesting for localized image editing tasks, \textit{i.e.} tasks that require modifying only a subset of pixels without altering the other ones.

\subsection{On reconstruction capabilities of Vector Quantization encoding  \label{subsection:localVQ}}
\captionsetup[subfigure]{labelformat=empty}
\begin{figure*}[t]
\centering
    \subfloat{\includegraphics[width=0.45\linewidth]{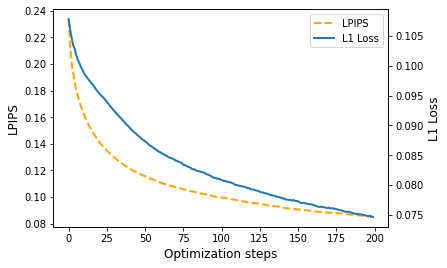}} \hspace{0.1cm}
    \subfloat{\includegraphics[width=0.38\linewidth]{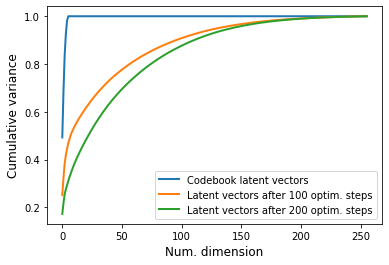}} \hspace{0.1cm}
    
    \caption{Analysis of reconstruction capabilities of VQGAN. On the left, we see the evolution of L1 loss and LPIPS between original and reconstructed image, when optimizing LPIPS over the latent vectors of VQGAN. On the right, we see the evolution of the number of dimension spanned by the latent vectors, before and after optimization.   \label{fig:vqgan_latent_vectors}}
\end{figure*} 

A limit of the framework for the use of VQGANs for image editing resides in its reconstruction capabilities. Indeed, the vector quantization operation induces a loss of information. In this section, we highlight some limits of the reconstructions of VQGANs and propose a simple optimization scheme to improve them.

In Figure \ref{fig:vqgan_latent_vectors}, we show that VQGAN struggles to properly reconstruct high-frequency details, for example complex backgrounds on FFHQ dataset \citep{karras2018style}. However, a simple optimization procedure over the latent space vectors improve the reconstruction quality of VQGANs and allows to recover fine details. The optimization consists in optimizing the LPIPS \citep{zhang2018unreasonable} over the latent vectors: $s^\star = \argmin_s \text{LPIPS}(D(s),I)$, where $I$ is the target image, $s$ the sequence of latent vectors and $D$ the decoder. We approximate the optimal latent vectors by gradient descent and initialize  $s$ as $s  = E(I) \in \mathbb{R}^{l \times  d}$. 
Furthermore, in Figure, we show a potential explanation of the limited reconstruction capabilities of VQGAN: the latent vectors of the codebook have a very low rank. After the optimization procedure, the latent vectors span more dimensions of the embedding space. 

\captionsetup[subfigure]{labelformat=empty}
\begin{figure*}
\centering
    \subfloat{\includegraphics[width=0.18\linewidth]{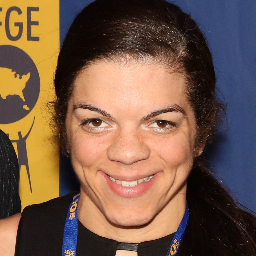}} \hspace{0.1cm}
    \subfloat{\includegraphics[width=0.18\linewidth]{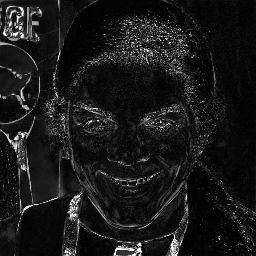}} \hspace{0.1cm}
    \subfloat{\includegraphics[width=0.18\linewidth]{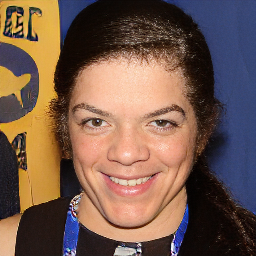}} \hspace{0.1cm}
    \subfloat{\includegraphics[width=0.18\linewidth]{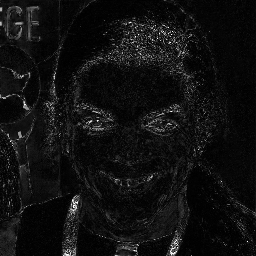}} \hspace{0.1cm}
    \subfloat{\includegraphics[width=0.18\linewidth]{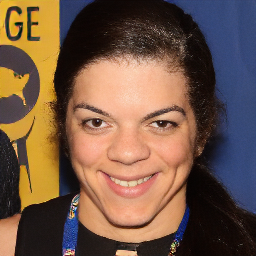}} \hspace{0.1cm}
    \\ \vspace{-0.25cm}
    \subfloat[Original.]{\includegraphics[width=0.18\linewidth]{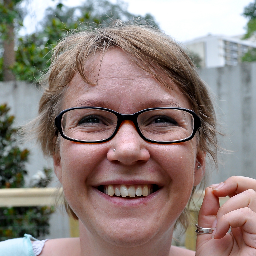}} \hspace{0.1cm}
    \subfloat[Difference between original and VQGAN reconstruction.]{\includegraphics[width=0.18\linewidth]{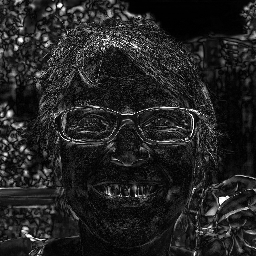}} \hspace{0.1cm}
    \subfloat[VQGAN reconstruction.]{\includegraphics[width=0.18\linewidth]{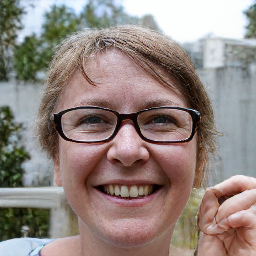}} \hspace{0.1cm}
    \subfloat[Difference between original and VQGAN plus latent optimization reconstruction.]{\includegraphics[width=0.18\linewidth]{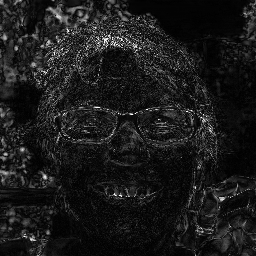}} \hspace{0.1cm}
    \subfloat[VQGAN plus latent optimization reconstruction.]{\includegraphics[width=0.18\linewidth]{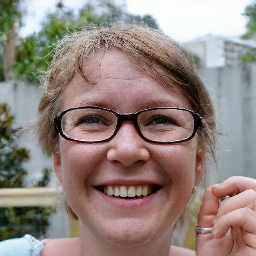}} \hspace{0.1cm}
    \\ \vspace{-0.25cm}
    \caption{Reconstructions from VQGAN on FFHQ, before and after optimization on latent vectors. \label{fig:vqgan_reconstruction}}
\end{figure*}

\section{Image editing with EdiBERT}
In this section, we describe the use of EdiBERT to perform image editing. Specifically, we distinguish three editing tasks: image denoising, image inpainting, and image composition (including image crossover, image compositing, and scribble-based editing). On a given dataset, we tackle each different task with the same pre-trained EdiBERT model and provide below a detailed explanation of the associated sampling algorithms. We run EdiBERT on FFHQ \citep{karras2018style} and LSUN Bedroom \citep{yu15lsun} at 256$\times$256. All metrics are reported on test-set of EdiBERT.

\textbf{Baselines. }
For each task, we run comparisons with baselines and state-of-the-art models based on GANs inversion methods. On FFHQ, we compare to ImageStyleGAN2++ \citep{abdal2020image2stylegan++} on pre-trained StyleGANs: StyleGAN2 \citep{karras2020analyzing} and StyleGAN2-ADA \citep{karras2020training}. Besides, we  run the solution proposed by \citet{chai2021using} where a StyleGAN2 model is inverted using a trained encoder. Finally, we use In-Domain GAN \citep{zhu2020domain}, a hybrid method combining an encoder with an optimization procedure minimizing reconstruction losses. We also compare to Co-Modulated GANs \citep{zhao2020large}, a conditional GAN for inpainting. 

\textbf{Metrics. } We follow the work of \citet{chai2021using} and use metrics assessing both fidelity and distribution fitting. The masked L1 metric \citep{chai2021using} measures the closeness between the generated image and the source image outside the edited areas. The density/coverage metrics \citep{naeem2020reliable} are robust versions of precision/recall metrics. Intuitively, density measures fidelity while coverage measures diversity. Finally, the FID \citep{heusel2017GANs} quantifies the distance between generated and target distributions. Moreover, we perform a user study on FFHQ image compositing. More details and quantitative results on LSUN Bedroom are presented in Appendix.

\subsection{Localized image denoising}
\captionsetup[subfigure]{labelformat=empty}
\begin{figure*}
\centering
    \subfloat{\includegraphics[width=0.49\linewidth]{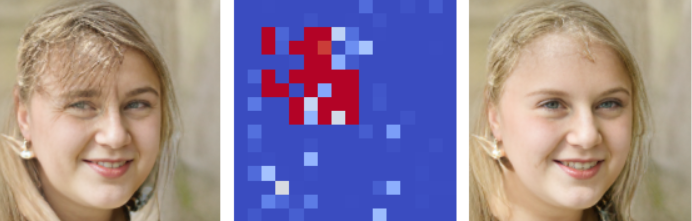}}
    \hspace{0.10cm}
    \subfloat{\includegraphics[width=0.49\linewidth]{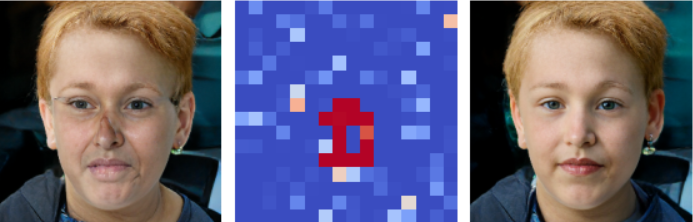}} \\
    \vspace{.12cm}
    \subfloat{\includegraphics[width=0.49\linewidth]{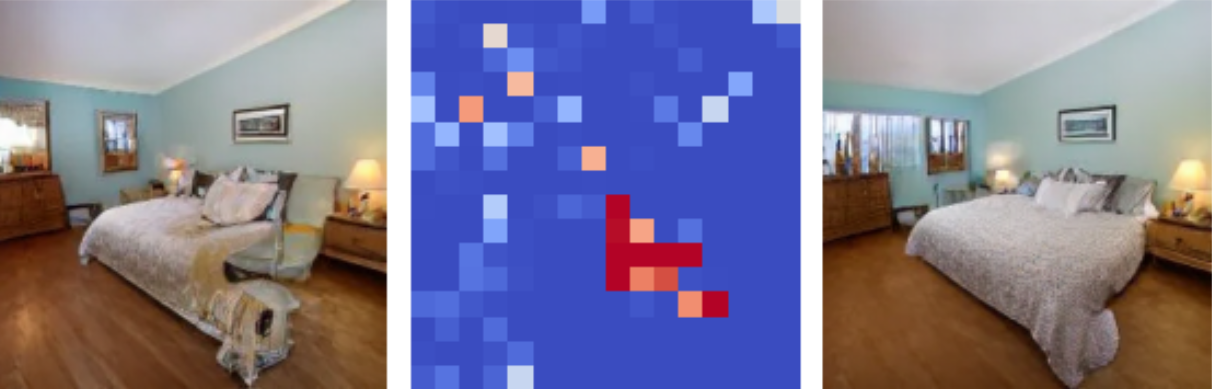}} 
    \hspace{0.10cm}
    \subfloat{\includegraphics[width=0.49\linewidth]{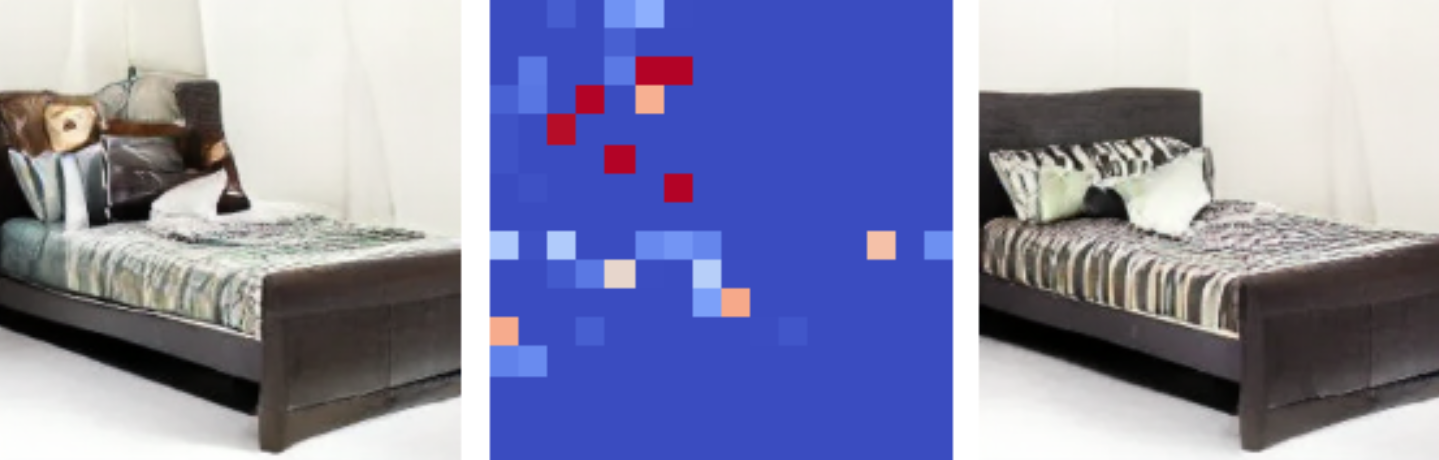}} \\
    \caption{Image denoising with EdiBERT: the color in the different heatmaps is proportional to the negative likelihood of the token. Tokens with lower likelihood appear in red in the heatmap and have a higher probability of being sampled to be edited. We observe that conditional distributions output by EdiBERT is a good way to detect anomalies and artifacts in the image space. \label{fig:image_denoising}}
\end{figure*}
Image denoising aims to improve the quality of a pre-generated image or improve a locally perturbed one. The model has to work without information on the localization of the perturbations. This means we need to find and replace the perturbed tokens with more likely ones to recover a realistic image. Thus, given a sequence $s=(s_1,\dots,s_L)$, we want to:
\begin{enumerate}
    \item Detect the tokens that do not fit properly in the sequence $s$.
    \item Change them for new tokens increasing the likelihood of the new sequence. 
\end{enumerate}

We desire a significantly more likely sequence with as few as possible token amendments. 
To do so, we measure the likelihood of each token $s_i$ based on the whole sequence $s$, aiming to compute $p(s_i | s)$, and replace the least-probable tokens considering them independently. That is, we propose to use the conditional probability output by the model in order to detect and sample the \textit{less likely} odd tokens. Some examples of image denoising are presented in Figure \ref{fig:image_denoising}, and we observe that EdiBERT is able to detect artifacts and replace them with more likely tokens. The full algorithm is given in the Algorithm \ref{algo:image_denoising}. 
\begin{algorithm}[ht]
\SetAlgoLined
 \textbf{Requires:} Sequence $(s_1, \hdots, s_l)$, BERT model $p_\theta$, number of iterations $T$\;
 \For{iterations in [0,T]}{
  Compute $p_i = \text{logit}(- p_\theta^i(s_i|s)), \forall i \in [1,l]$ \;
  Sample $p \sim (p_1, \hdots, p_l)$ (\textit{less likely position})\;
  Sample $t \in Z \sim p_\theta^p(\cdot|s)$ \;
  Insert sampled token: $s_i \gets t$ \;
 }
 $\text{Image} \gets \text{Decoder}(s)$\;
 \KwResult{Image}
 \caption{Image denoising \label{algo:image_denoising}}
\end{algorithm}

\subsection{Image inpainting}
In this setting, we have access to a masked image $i_m \in \mathbb{R}^{h \times w \times c}$ along with the location of the binary mask $m \in \mathbb{R}^{h \times w}$. $i_m$ has been obtained by masking an image $i \in \mathbb{R}^{h \times w \times c}$ as follows: $i_m = i \odot m$ with $\odot$  pointwise multiplication. The goal of image inpainting is to generate an image $\hat{i}$ that is both realistic (high density) and conserves non-masked parts, that is $\hat{i} \odot (1-m) = i \odot (1-m)$. 

\begin{algorithm}[th]
\SetAlgoLined
 \textbf{Requires:} Masked (or edited) image $i_m$, mask $m$, Encoder E, Decoder D, BERT model $p_\theta$, epochs e, periodic  collage c, optimization steps optim\_steps\;
 $s \gets Q(E(i_m))$\;
 $\text{latent\_mask} \gets \text{get\_mask\_in\_latent\_space}(m)$\;
 \If{task is inpainting}{
 $s \gets s \times \text{latent\_mask} + \text{rand} \times (1-\text{latent\_mask})$\; }
 \For{e in [0,epochs]}{
 \For{p in $\text{chosen\_order}(\text{latent\_mask})$}{
  Sample token $t \in Z \sim p_\theta^p(\cdot|s)$ \;
  Insert sampled token: $s_p \gets t$ \;
  \If{p\%c=0 \text{(collage)}}{
    Encode image post-collage: $s \gets E(i_m \odot m + D(s) \odot (1-m))$\;}
 }}
 $s^0 \gets s$ \;
 \For{i in [0, optim\_steps]:}{
 $L = L_p \big( (D(s) - i_m) \odot m \big) + L_p \big( (D(s) - D(s^0) ) \odot (1-m) \big)$ \;
 $s \gets s + \epsilon * \text{Adam}(\nabla_s L, s)$ \;
 }
 $\text{Image} \gets \text{Decoder}(s)$\;
 \KwResult{Image}
 \caption{Image inpainting/composition \label{algo:image_inpainting}}
\end{algorithm}

\begin{figure}
\centering
    \subfloat{\includegraphics[width=0.142\linewidth]{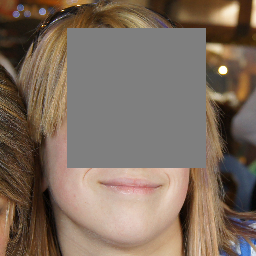}}
    \subfloat{\includegraphics[width=0.142\linewidth]{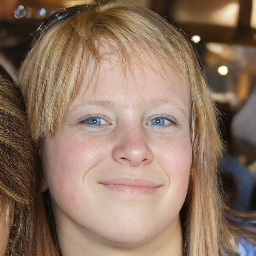}}
    \subfloat{\includegraphics[width=0.142\linewidth]{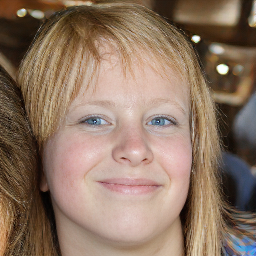}} 
    \subfloat{\includegraphics[width=0.142\linewidth]{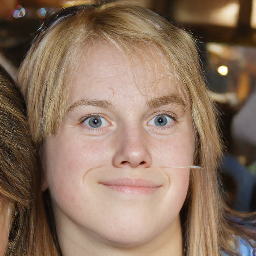}}
    \subfloat{\includegraphics[width=0.142\linewidth]{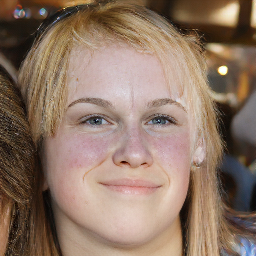}}
    \subfloat{\includegraphics[width=0.142\linewidth]{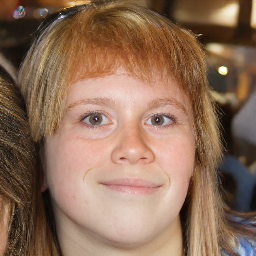}}
    \subfloat{\includegraphics[width=0.142\linewidth]{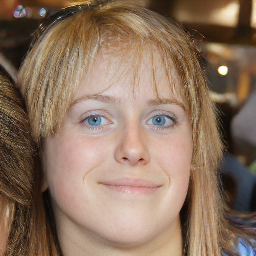}}\\ \vspace{-0.3cm}
    \subfloat{\includegraphics[width=0.142\linewidth]{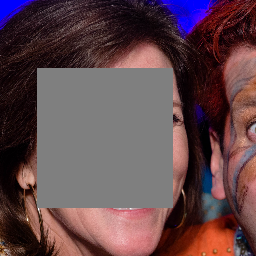}}
    \subfloat{\includegraphics[width=0.142\linewidth]{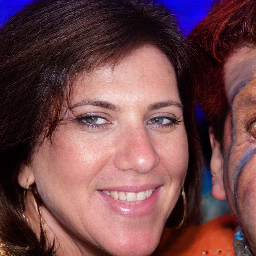}}
    \subfloat{\includegraphics[width=0.142\linewidth]{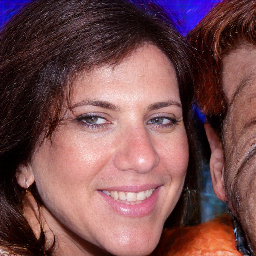}} 
    \subfloat{\includegraphics[width=0.142\linewidth]{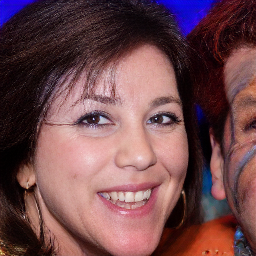}}
    \subfloat{\includegraphics[width=0.142\linewidth]{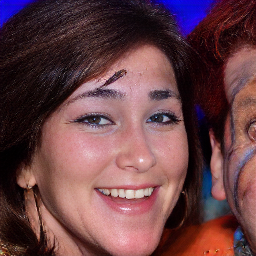}}
    \subfloat{\includegraphics[width=0.142\linewidth]{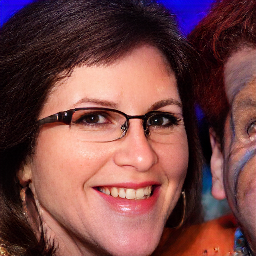}}
    \subfloat{\includegraphics[width=0.142\linewidth]{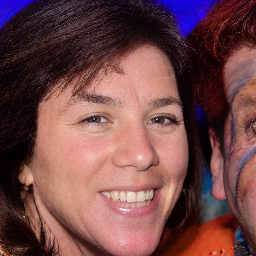}}\\  \vspace{-0.3cm}
    \subfloat{\includegraphics[width=0.142\linewidth]{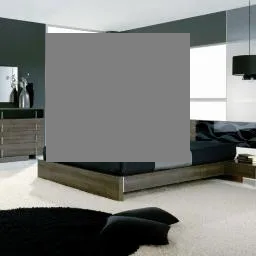}}
    \subfloat{\includegraphics[width=0.142\linewidth]{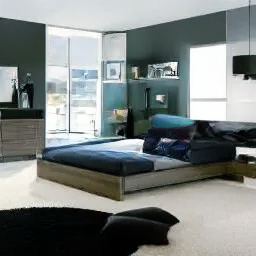}}
    \subfloat{\includegraphics[width=0.142\linewidth]{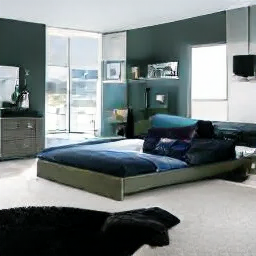}} 
    \subfloat{\includegraphics[width=0.142\linewidth]{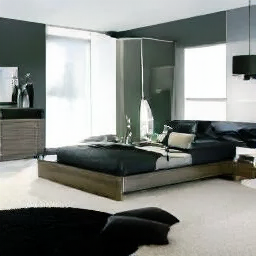}}
    \subfloat{\includegraphics[width=0.142\linewidth]{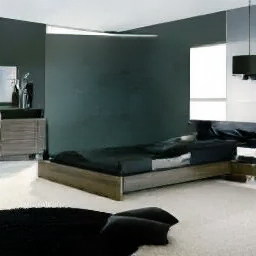}}
    \subfloat{\includegraphics[width=0.142\linewidth]{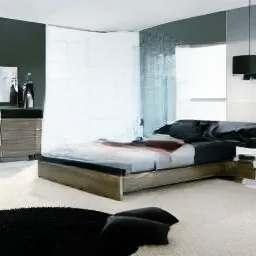}}
    \subfloat{\includegraphics[width=0.142\linewidth]{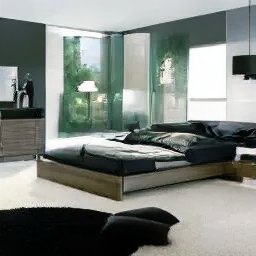}}\\  \vspace{-0.3cm}
    \subfloat[Masked]{\includegraphics[width=0.142\linewidth]{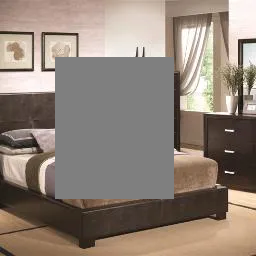}}
    \subfloat[EdiBERT] {\includegraphics[width=0.142\linewidth]{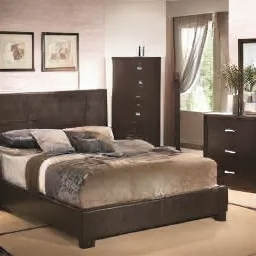}}
    \subfloat[ (a)] {\includegraphics[width=0.142\linewidth]{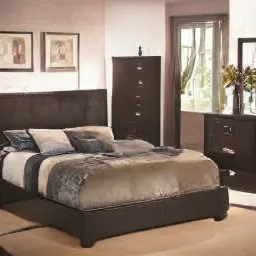}} \subfloat[(b)]{\includegraphics[width=0.142\linewidth]{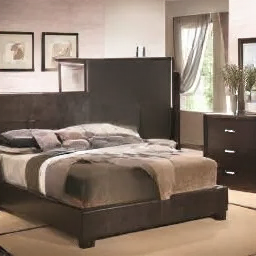}}
    \subfloat[(c)] {\includegraphics[width=0.142\linewidth]{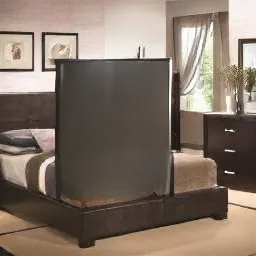}}
    \subfloat[(d)] {\includegraphics[width=0.142\linewidth]{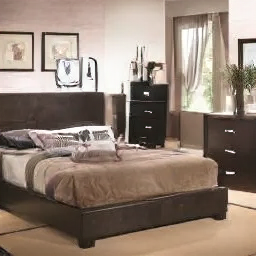}} 
    \subfloat[(e)] {\includegraphics[width=0.142\linewidth]{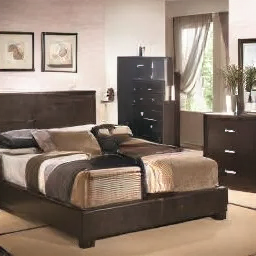}} \\ 
    \caption{Ablation study for inpainting. 
    Components removed are (a) optimization, (b) dilation, (c) randomization, (d) collage, (e) spiraling (random order instead).
    \label{fig:image_completion_results}
    } \vspace{-0.4cm}
\end{figure}

Among the different tasks in image manipulation, image inpainting stands out. Indeed, when masking a specific area of an image, one shall not consider the pixels within the mask to recover the target image. Interestingly, this is the main difference with the scribble-based or image compositing editing tasks, where, on the opposite, one must consider the modifications brought by user input.
Due to these peculiarities, the image inpainting task requires specific care to reach a state-of-the-art performance; this is why we introduce five different elements to our approach. We validate these elements with visual results in Figure \ref{fig:image_completion_results} and and an ablation study in Table \ref{table:ablation_inpainting}.

\begin{enumerate}
    \item \textbf{Erasing mask influence with random tokens (randomization).} To delete the information contained in the mask, all the tokens within the mask are given random values. 
    \item \textbf{Dilating the mask in the token space (dilation).} As shown in Figure \ref{fig:vqgan_encoding}, some tokens outside of the down-sampled mask in the latent space are also impacted by the mask. Consequently, only modifying tokens inside the down-sampled mask might not be enough and could lead to images with irregularities on the borders. As a solution, we propose to apply a dilation on the down-sampled mask and show an example of this in Figure \ref{fig:image_completion_results}. We observe that dilating the mask helps to better blend the target image's completion since the boundaries are removed.
    \item \textbf{Defining a new ordering of positions (random ordering).} There is no pre-defined ordering of positions in EdiBERT. One can thus look for optimal sampling of positions. We argue that by sampling positions randomly, one does not fully leverage the spatial location of the mask. Instead, we propose to use a spiral ordering that goes from the border of the mask to the inside. Figure \ref{fig:image_completion_results} shows examples of image completion with this pre-defined ordering or random ordering. Results from Table \ref{table:ablation_inpainting} confirm the advantage of this ordering. 
    \item \textbf{Periodic image collage (collage).} To preserve fidelity to the original image, we periodically perform an image collage between the masked image and the image decoded from the current latent representation. Without this collage trick, the reconstruction can diverge too much from the input image. When performing optimization without collage, we observe border effects (see Row 3, last column in Figure \ref{fig:image_completion_results}).
    \item \textbf{Online optimization on latent sequence (optimization). } To further improve fidelity to the masked image $i_m$, the final stage of our algorithm consists in an optimization procedure on the latent sequence $s \in \mathbb{R}^{h \times w \times d}$. The objective function is defined as:
\begin{equation}
    L = L_p \big( (D(s) - i_m) \odot m \big) + L_p \big( (D(s) - D(s^0) ) \odot (1-m) \big)
\end{equation}
where $L_p$ is a perceptual loss \citep{zhang2018unreasonable}, and $s^0$ is the initial sequence from EdiBERT. Intuitively, the first term enforces the decoded image to get closer to the masked image $i_m$, while the second term is a regularization enforcing the decoded image to stay similar to the completion proposed by the transformer's likelihood.

If this loss function is optimized over the image space, the resulting image would be the following one: $i_m \odot m + D(s^0) \odot (1-m)$. The resulting image would be non-smooth, with visible border effects. However, when optimizing over the sequence of latent vectors $s$, the decoder regularizes the optimization and avoids non-smooth borders. We illustrate in Figure \ref{fig:vqgan_latent_vectors} and Figure \ref{fig:image_completion_results} that the optimization leads to a better-preserved source image.
\end{enumerate}

\setlength{\tabcolsep}{0.1pt}
\begin{table}[t]
\caption{Image inpainting and compositing on FFHQ $256\times256$. Com-GAN is a model specific for image inpainting, ID-GAN handles several editing tasks but not inpainting, while other methods handle both. We remove I2SG++ from the user study, since  I2SG$^\dagger$++ is the same method with a better GAN backbone, \textit{i.e.} StyleGAN2-ADA \citep{karras2020training}.
\textbf{Bold: $1^{st}$ rank}, {\color{blue}blue: $2^{nd}$ rank}.}
\label{main_table}
\centering
\noindent
\begin{tabular}{N N N N N N N N}\toprule
\multicolumn{1}{c }{\textbf{}} & \multicolumn{4}{c }{\textbf{Inpainting}} & \multicolumn{3}{c }{\textbf{Compositing}} \\ 
\cmidrule(lr){2-5}
\cmidrule(ll){6-8}
 & \textbf{Masked L1 $\downarrow$} & \textbf{FID $\downarrow$} & \textbf{Dens. $\uparrow$} & \textbf{Cover. $\uparrow$} & \textbf{Masked L1 $\downarrow$} & \textbf{Dens. $\uparrow$} & \textbf{User study $\uparrow$} \\ 
\cmidrule(lr){1-1}
\cmidrule(lr){2-5}
\cmidrule(ll){6-8}
\multicolumn{1}{ c }{I2SG++ \citep{abdal2020image2stylegan++}} & 0.0767 & 23.6 & 0.99 & 0.88 & 0.0851 & 0.77 & - \\ 
\multicolumn{1}{ c }{I2SG$^\dagger$++ \cite{abdal2020image2stylegan++}} &  0.0763 & 22.1 & {\color{blue}1.25} & 0.91 & 0.0866 & \textbf{1.07} & 8.3\%  \\ 
\multicolumn{1}{ c }{LC \citep{chai2021using}} & 0.1027 & 27.9 & 1.12 & 0.84 & 0.1116 & {\color{blue}1.00}  &  14.8\% \\ 
\multicolumn{1}{ c }{EdiBERT $\star$} & {\color{blue}0.0290} & {\color{blue}13.8} & {1.16} & {\color{blue}0.98} & \textbf{0.0307} & 0.94 & \textbf{61.2}\% \\ \midrule
\multicolumn{1}{ c }{Com-GAN \citep{zhu2020domain}} & \textbf{0.0086} &  \textbf{10.3} & \textbf{1.42} & \textbf{1.00} & - & - & - \\ 
\multicolumn{1}{ c }{ID-GAN \citep{zhu2020domain}} & - & - & - & - & {\color{blue}0.0570}  & 0.75 & {\color{blue}15.7}\% \\ 
\bottomrule
\end{tabular}
\end{table}

\captionsetup[subfigure]{labelformat=empty}

\paragraph{Analyzing the results:} Observing the results in Table \ref{main_table}, we see that the specialized method com-GAN \citep{zhao2020large} outperforms non-specialized methods on image inpainting. This was expected since it is the only method that has been trained specifically for this task. Note that the trained model co-mod GAN cannot be used in any other image manipulation task. Compared with the non-specialized method, EdiBERT always provides better fidelity to the source image (lower Masked L1) and  realism (best FID and top-2 density). 

\subsection{Image composition}
In this setting, we have access to a non-realistically edited image $i_e \in \mathbb{R}^{h \times w \times c}$. The edited image $i_e$ is obtained by a composition between a source image $i_s \in \mathbb{R}^{h \times w \times c}$ and a target image $i_t \in \mathbb{R}^{h \times w \times c}$. The target image can be a user-drawn scribble or another real image in the case of image compositing. Besides, pixels are edited inside a binary mask $m \in \mathbb{R}^{h \times w}$, which indicates the areas modified by the user. Thus, the final edited image is computed pointwise as: 
\begin{equation}
    i_e = i_s \odot m + i_t \odot (1-m). 
\end{equation}
Image composition aims to transform an edited image $i_e$ to make it more realistic and faithful without limiting the changes outside the mask. 
We note the source image $i_s$ outside the mask and the edits of the target image $i_m$ for the inserted elements in the edition mask. Three tasks fall under this umbrella: \textit{scribble-based editing}, \textit{image compositing}, and \textit{image crossovers}.


Results of image compositing on FFHQ are presented in Table \ref{main_table}. EdiBERT always has the lowest masked L1. We also present the results from a user study in Table \ref{main_table}. 30 users were shown 40 original and edited images, along with four results (EdiBERT and baselines). They were asked which one is preferable, accounting for both fidelity and realism. The survey shows users on average prefer EdiBERT over competing approaches. We give the detailed answers of the user study in Appendix.

\captionsetup[subfigure]{labelformat=empty}
\begin{figure*}
\centering
    \subfloat[]{\includegraphics[width=0.18 \linewidth]{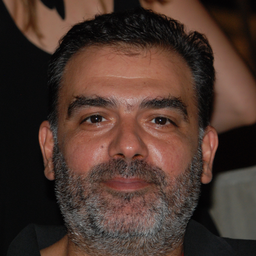}}
    \subfloat[]{\includegraphics[width=0.18 \linewidth]{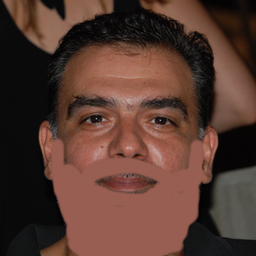}}
    \subfloat[]{\includegraphics[width=0.18 \linewidth]{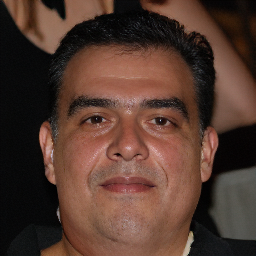}}
    \subfloat[]{\includegraphics[width=0.18 \linewidth]{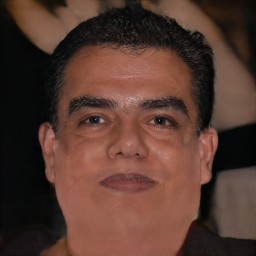}}
    \subfloat[]{\includegraphics[width=0.18 \linewidth]{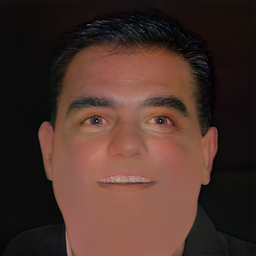}}
    \\  \vspace{-0.8cm}
    \subfloat[]{\includegraphics[width=0.18 \linewidth]{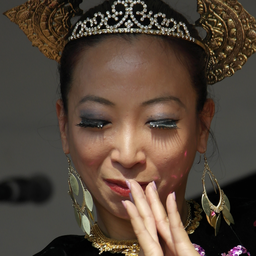}}
    \subfloat[]{\includegraphics[width=0.18 \linewidth]{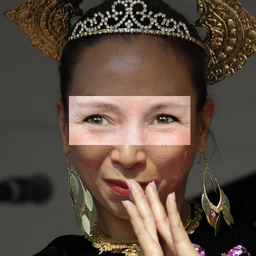}}
    \subfloat[]{\includegraphics[width=0.18 \linewidth]{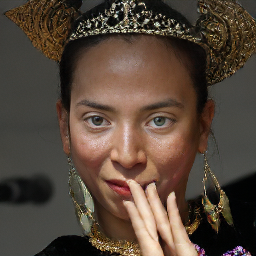}}
    \subfloat[]{\includegraphics[width=0.18 \linewidth]{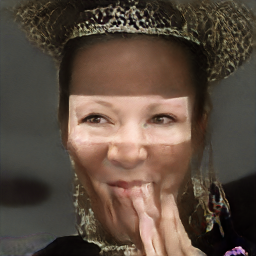}}
    \subfloat[]{\includegraphics[width=0.18 \linewidth]{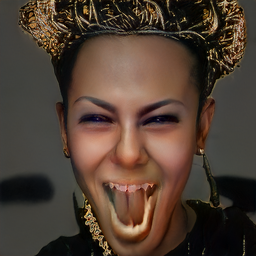}} \\ \vspace{-0.8cm}
    \subfloat[Source]{\includegraphics[width=0.18 \linewidth]{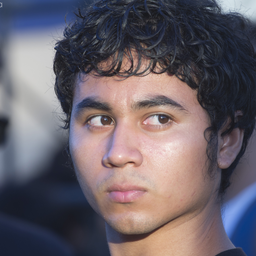}}
    \subfloat[Composite]{\includegraphics[width=0.18 \linewidth]{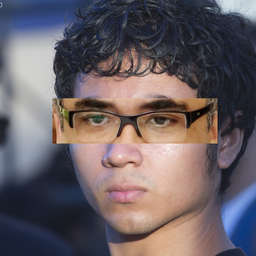}}
    \subfloat[EdiBERT $(\star)$]{\includegraphics[width=0.18 \linewidth]{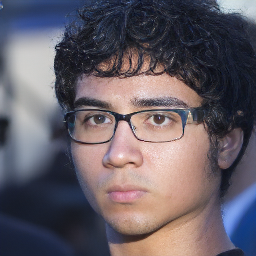}}
    \subfloat[ID-GAN \cite{zhu2020domain}]{\includegraphics[width=0.18 \linewidth]{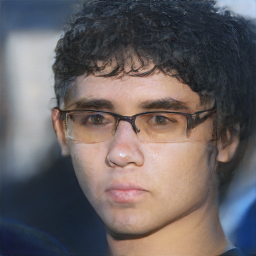}}
    \subfloat[I2SG$^\dagger$++ \cite{abdal2020image2stylegan++,karras2020training}]{\includegraphics[width=0.18 \linewidth]{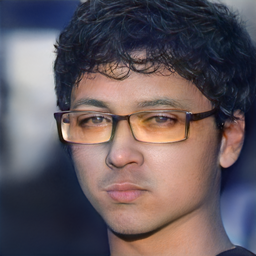}}
    \\
    \subfloat[]{\includegraphics[width=0.18 \linewidth ]{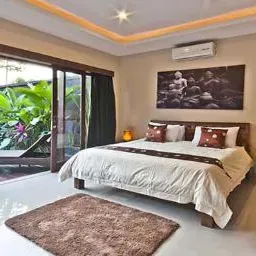}}
    \subfloat[]{\includegraphics[width=0.18 \linewidth ]{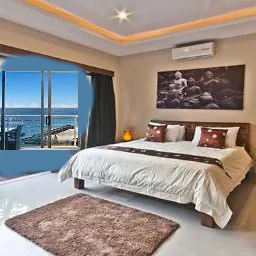}}
    \subfloat[]{\includegraphics[width=0.18 \linewidth ]{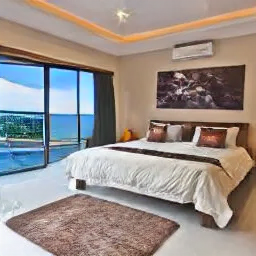}}
    \subfloat[]{\includegraphics[width=0.18 \linewidth ]{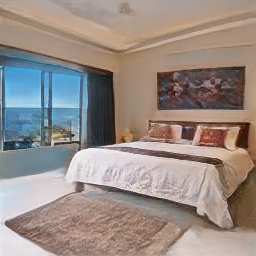}}
    \subfloat[]{\includegraphics[width=0.18 \linewidth ]{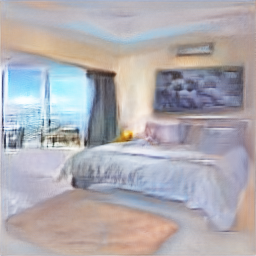}}
    \\ \vspace{-0.8cm}
    \subfloat[]{\includegraphics[width=0.18 \linewidth ]{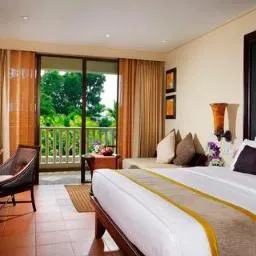}}
    \subfloat[]{\includegraphics[width=0.18 \linewidth ]{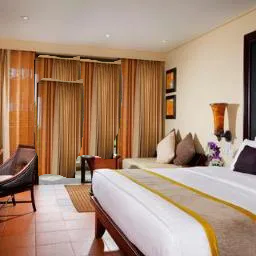}}
    \subfloat[]{\includegraphics[width=0.18 \linewidth ]{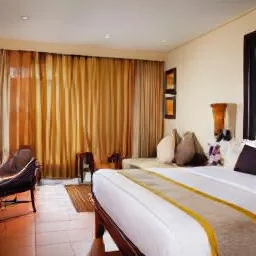}}
    \subfloat[]{\includegraphics[width=0.18 \linewidth ]{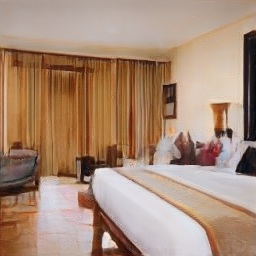}}
    \subfloat[]{\includegraphics[width=0.18 \linewidth ]{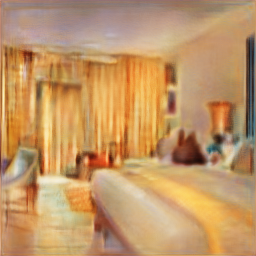}}
    \\ \vspace{-0.8cm}
    \subfloat[Source]{\includegraphics[width=0.18 \linewidth ]{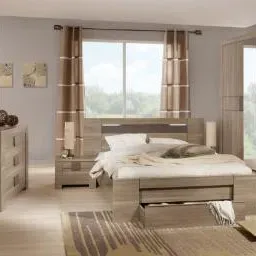}}
    \subfloat[Composite]{\includegraphics[width=0.18 \linewidth ]{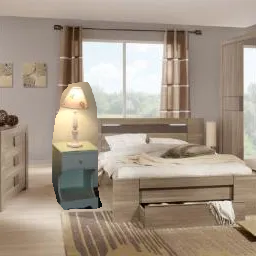}}
    \subfloat[EdiBERT $(\star)$]{\includegraphics[width=0.18 \linewidth ]{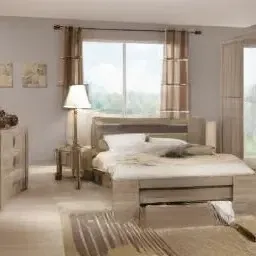}}
    \subfloat[ID-GAN \cite{zhu2020domain}]{\includegraphics[width=0.18 \linewidth ]{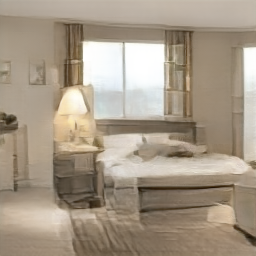}}
    \subfloat[I2SG \cite{abdal2019image2stylegan}]{\includegraphics[width=0.18 \linewidth ]{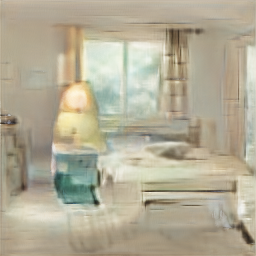}}
    \caption{Scribble-based editing and image compositing: EdiBERT preserves more the connection to the source image while being able to fit the inserted object properly. This confirms the quantitative results in Table \ref{main_table} where EdiBERT appears to be leading in fidelity and realism. \label{fig:image_composition}}
\end{figure*}

\setlength{\tabcolsep}{0.4pt}
\begin{table}[h]
\caption{Inpainting: Ablation study on the components of EdiBERT sampling algorithm. EdiBERT (1st row) shows the best tradeoff between fidelity (masked L1) and quality (FID, density/coverage).  \textbf{Bold: $1^{st}$ rank}, {\color{blue}blue: $2^{nd}$ rank}.  \label{table:ablation_inpainting}}
\centering
\begin{tabular}{N N N N N N N N N}\toprule
Ordering & Optim- & Random- & Collage & Dilation & Masked & FID & Density & Coverage\\
 & ization & ization & & & L1 $\downarrow$ & $\downarrow$ & $\uparrow$ & $\uparrow$ \\ \hline
 Spiral & \text{\checkmark} & \text{\checkmark} & \text{\checkmark} & \text{\checkmark} & 0.0201 & \textbf{19.4} & 1.14 & \textbf{0.96} \\
 Random & \text{\checkmark} & \text{\checkmark} & \text{\checkmark} & \text{\checkmark} & 0.0206 & 20.7 & 1.13 & {\color{blue}0.95} \\
   Spiral & X & \text{\checkmark} & \text{\checkmark} & \text{\checkmark} & 0.0299 & 20.3 & {\color{blue}1.20} & 0.94 \\
  Spiral & \text{\checkmark} & X & \text{\checkmark} & \text{\checkmark} & {\color{blue}0.0198} & 20.5 & \textbf{1.26} & 0.92  \\ 
   Spiral & \text{\checkmark} & \text{\checkmark} & X & \text{\checkmark} & 0.0252 & {\color{blue}19.9} & 1.11 & 0.95 \\
  Spiral & \text{\checkmark} & \text{\checkmark} & \text{\checkmark} & X & \textbf{0.0197} & 23.3 & 0.96 & 0.91 \\ \bottomrule
\end{tabular}
\end{table}

\section{Discussions}
EdiBERT is a bidirectional transformers model that can tackle multiple editing tasks from one single training. One of the key elements of the proposed method is that it does not require having access to paired datasets (source, target), or unpaired image datasets. This property shows how flexible EdiBERT is and why it can be easily applied to different tasks. Overall, the proposed framework is simple and tractable: 1) train a VQGAN \cite{esser2021taming}, 2) train an EdiBERT model following the objective defined in \eqref{eq:bidirectionnal}. 

Interestingly, for simple applications, one can directly train EdiBERT based on the representations output by the VQGAN pre-trained on ImageNet. However, for more complex data or when dealing with multiple domains, one might have to train a specialized codebook, which requires a large auto-encoder and a lot of data. Another EdiBERT's drawback is related to the core interest of image editing. Since the tokens are predominantly localized, EdiBERT is perfectly suited for small manipulations that only require amending a few numbers of tokens. However, some manipulations such as zooms or rotations require changing large areas of the source image. In these cases, modifying a large number of tokens might be  more demanding. 

Similarly to other image generative models, EdiBERT might also be used to create and propagate fake beliefs via deepfakes, as discussed in \cite{fallis2020epistemic}.

\section{Conclusion}
In this paper, we demonstrated the possibility to perform several editing tasks by using the same pre-trained model. The proposed framework is simple and aims at making a step towards a unified model able to do any conceivable manipulation task on images. An exciting direction of research would be to extend the editing capabilities of EdiBERT to global transformations (\textit{e.g.} zoom, rotation).

\clearpage
\bibliography{main}
\bibliographystyle{tmlr}

\appendix
\begin{Large}\textbf{Appendix}
\end{Large}

\section{Implementation details}
The code for the implementation of EdiBERT is available on GitHub at the following link \href{https://github.com/EdiBERT4ImageManipulation/EdiBERT}{https://github.com/EdiBERT4ImageManipulation/EdiBERT}.

A pre-trained model on FFHQ is available on a linked Google Drive. Notebooks to showcase the model have also been developped. 

\subsection{Training hyper-parameters}
We use the same architecture than \cite{esser2021taming} for both VQGAN and transformer. On LSUN Bedroom and FFHQ, we use a codebook size of 1024. For the transformer, we use a model with 32 layers of width 1024.

To train the transformer with 2D masking strategy, we generate random rectangles before flattening $Q(E(I))$. The height of rectangles is drawn uniformly from $[0.2\times h,0.5 \times h]$. Similarly, the width of rectangles is drawn uniformly from $[0.2\times w,0.5 \times w]$. In our experiments, since we work at resolution $256 \times 256$ and follow the downsampling factor of 4 from \cite{esser2021taming}, we have $h=w=256/4=16$. 

Tokens outside the rectangle are used as input, to give context to the transformer, but not for back-propagation. Tokens inside the rectangle are used for back-propagation. $p_{\text{rand}}=90\%$ of tokens inside the mask are put to random tokens, while $p_{\text{same}}=1-p_{\text{rand}} = 10\%$ are given their initial value. Although we did not perform a large hyper-parameter study on this parameter, we feel it is an important one. The lower $p_{\text{rand}}$, the more the learned distributions $p^i_\theta(.|s)$ will be biased towards the observed token $s_i$. However, setting $p_{\text{rand}}=1$ leads to a model that diverges too fast from the observed sequence. 

\subsection{Inference hyper-parameters}
\subsubsection{Image inpainting.}
We set the number of epochs to 2, collage frequency to 4 per epoch, top-k sampling to 100, dilation to 1, and number of optimization steps to 200. We apply a gaussian filter on the mask for the periodic image collage.

Additionally, we use these two implementation details. 1) We use two latent masks: the latent down-sampled mask $\text{latent\_mask}_1$, and the dilated mask $\text{latent\_mask}_2$, obtained by a dilation of $\text{latent\_mask}_1$. The randomization is done with $\text{latent\_mask}_1$, such that no information from the unmasked parts of the image is erased. However, the selection of positions that are re-sampled by EdiBERT is done with $\text{latent\_mask}_2$. 2) At the second epoch, we randomize the token value, at the position that is being replaced. This is only done for image inpainting.

\subsubsection{Image compositing.}
We set the number of epochs to 2, collage frequency to 4 per epoch, top-k sampling to 100, dilation to 1, and number of optimization steps to 200. We apply a gaussian filter on the mask for the periodic image collage. Contrarily to inpainting, we do not randomize such that EdiBERT samples stay closer to the original sequence. 

\section{Additional experimental results}
We give additional comparisons on FFHQ and LSUN Bedroom, for the following tasks: image inpainting in Table \ref{table:image_inpainting_results}, image crossovers in Table \ref{table:crossovers}, and image composition in Table \ref{table:editing}. All these experiments are run on the test-set of EdiBERT.  Note that concurrent methods based on StyleGAN2 were trained on the full dataset, which advantages them. 

\textbf{Inpainting. } We use 2500 images. On FFHQ, we provide results for free-form masks and rectangular masks. The height of rectangular masks is drawn uniformly from $[0.4\times h,0.6 \times h]$ with $h=256$, and similarly for the width. For non-rectangular masks generations, we follow the procedure of \cite{chai2021using}: we draw a binary mask at low-resolution $6\times 6$ and uspsample it to $256\times 256$ with bilinear interpolation.

The ablation study in Table \ref{table:ablation_inpainting} of main paper is performed on free-form masks. Results in Table \ref{main_table} of main paper are on rectangular masks. On LSUN Bedroom, we provide results for rectangular masks. 

\textbf{Crossovers. } We generate 2500 crossovers from random pairs of images, on both FFHQ and LSUN Bedroom.

\textbf{Editing/Compositing. } We create small datasets of 100 images from the test set of EdiBERT for FFHQ scribble-based editing, FFHQ  compositing and LSUN Bedroom compositing. A user study on FFHQ  compositing is presented in main paper with statistically significant number of votes. We also provide some metrics in \ref{table:editing}. Because of the small size of the dataset, we only report masked L1 and density. For density, the support of the real distribution is estimated with 2500 real points, and density is averaged over the individual density of the 100 generated images.  

\begin{table}[ht]
\centering
\begin{tabular}{l|N N N N}
 & Masked L1 $\downarrow$ & FID $\uparrow$ & Density $\uparrow$ & Coverage $\uparrow$ \\ \hline
\multicolumn{4}{l}{\textbf{FFHQ:} rect. masks} \\\hline
I2SG++ \cite{abdal2020image2stylegan++} & 0.0767 & 23.6 & 0.99 & 0.88\\
I2SG$^\dagger$++ \cite{abdal2020image2stylegan++,karras2020training} & 0.0763 & 22.1 & \textbf{1.25} & 0.91  \\
LC \cite{chai2021using} & 0.1027 & 27.9 & 1.12 & 0.84 \\
EdiBERT  ($\star$) & 0.0290 & 13.8 & 1.16 & 0.98 \\\hline
Co-mod. GAN \cite{zhao2020large}  & \textbf{0.0128} &  \textbf{4.7} & {1.24} & \textbf{0.99}   \\ \hline  
\multicolumn{4}{l}{\textbf{FFHQ:} free-form masks} \\\hline
I2SG++ \cite{abdal2020image2stylegan++}  & 0.0440 & 22.3 & 0.92 & 0.89 \\
I2SG$^\dagger$++ \cite{abdal2020image2stylegan++,karras2020training} & 0.0435 & 21.1 & 1.17 & 0.91 \\
LC \cite{chai2021using} & 0.0620 & 27.9 & 1.22 & 0.85 \\
EdiBERT ($\star$) & 0.0201 & 19.4 & 1.14 & 0.96  \\\hline
Com-GAN \cite{zhao2020large} & \textbf{0.0086} &  \textbf{10.3} & \textbf{1.42} & \textbf{1.00} \\ \hline 
\multicolumn{4}{l}{\textbf{LSUN Bedroom}: rect. masks} \\ \hline
I2SG \cite{abdal2019image2stylegan} & 0.1125 & 50.2 & 0.04& 0.04 \\
EdiBERT ($\star$) & \textbf{0.0288} & \textbf{11.3} & \textbf{0.89} & \textbf{0.97} \\\hline
\end{tabular}
\caption{Image inpainting. \label{table:image_inpainting_results}}
\end{table}

\begin{table}
\centering
\begin{tabular}{l|N N N N}
 & Masked L1 $\downarrow$ & FID $\uparrow$ & Density $\uparrow$ & Coverage $\uparrow$ \\ \hline
\multicolumn{4}{l}{\textbf{FFHQ}} \\\hline
I2SG++ \cite{abdal2020image2stylegan++}  & 0.1141 & 29.4 & 0.97 & 0.78 \\
I2SG$^\dagger$++ \cite{abdal2020image2stylegan++,karras2020training} & 0.1133 & 26.9 & \textbf{1.35} & 0.82 \\
ID-GAN \cite{zhu2020domain} & 0.0631 & 23.2 & 0.88 & 0.83 \\
LC \cite{chai2021using} & 0.1491 & 31.9 & 1.17 & 0.77\\
EdiBERT  ($\star$) & \textbf{0.0364} & \textbf{19.7} & {1.05} & \textbf{0.88} \\ \hline \hline 
\multicolumn{4}{l}{\textbf{LSUN Bedroom}} \\\hline
I2SG \cite{abdal2019image2stylegan} & 0.1123 & 45.7 & 0.12 & 0.20 \\
ID-GAN \cite{zhu2020domain} & 0.0682 & 21.4 & 0.35 & 0.57 \\
EdiBERT  ($\star$) & \textbf{0.0369} & \textbf{12.4} & \textbf{0.64} & \textbf{0.84} \\ \hline  
\end{tabular}
\caption{Image crossover. \label{table:crossovers}}
\end{table}

\begin{table}
\centering
\begin{tabular}{l|N N }
 & Masked L1 $\downarrow$  & Density $\uparrow$ \\ \hline
\multicolumn{2}{l}{\textbf{FFHQ scribble-edits}} \\ \hline
I2SG++ \cite{abdal2020image2stylegan++}  & 0.7811 & 0.91 \\
I2SG$^\dagger$++ \cite{abdal2020image2stylegan++,karras2020training} & 0.0777  & 1.11    \\
ID-GAN \cite{zhu2020domain} & 0.0461 &  0.79\\
LC \cite{chai2021using} & 0.1016 & \textbf{1.14}  \\
EdiBERT  ($\star$) & \textbf{0.0281} & 0.96 \\ \hline
\multicolumn{2}{l}{\textbf{FFHQ compositing}} \\\hline
I2SG++ \cite{abdal2020image2stylegan++}  & 0.0851 & 0.77   \\
I2SG$^\dagger$++ \cite{abdal2020image2stylegan++,karras2020training} & 0.0866  & \textbf{1.07}   \\
ID-GAN \cite{zhu2020domain} & 0.0570 & 0.75    \\
LC \cite{chai2021using} & 0.1116  & 1.00  \\
EdiBERT  ($\star$) & \textbf{0.0307} & 0.94 \\ 
\hline  
\multicolumn{2}{l}{\textbf{LSUN Bedroom compositing}} \\\hline
I2SG \cite{abdal2019image2stylegan} & 0.1285 & 0.25 \\
ID-GAN \cite{zhu2020domain}  & 0.0484  & 1.45 \\
EdiBERT  ($\star$) & \textbf{0.0247} & \textbf{1.49}  \\ \hline  
\end{tabular}
\caption{Image editing. \label{table:editing}}
\end{table}

\clearpage
\section{Baselines}
We use the implementation and pre-trained models from the following repositories.

ID-GAN \citep{zhu2020domain}: \href{https://github.com/genforce/idinvert\_pytorch}{https://github.com/genforce/idinvert\_pytorch}, which has pre-trained models on FFHQ 256x256 and LSUN Bedroom 256x256.

I2SG++ and I2SG$\dagger$++\citep{karras2020analyzing,karras2020training,abdal2020image2stylegan++}: \\ \href{https://github.com/NVlabs/stylegan2-ada-pytorch}{https://github.com/NVlabs/stylegan2-ada-pytorch}. We tested projections with the following pre-trained models on FFHQ: StyleGAN2 \citep{karras2020analyzing} at resolution 256x256, and StyleGAN2-Ada \citep{karras2020training} at resolution FFHQ 1024x1024. For evaluation, we downsample the 1024x1024 generated images to 256x256.

LC \citep{chai2021using}:  \href{https://github.com/chail/latent-composition}{https://github.com/chail/latent-composition}. We use the pre-trained encoder and StyleGAN2 generator, for FFHQ at resolution 1024x1024. For evaluation, we downsample the 1024x1024 generated images to 256x256.

Com-GAN \cite{zhao2020large}: \href{https://github.com/zsyzzsoft/co-mod-gan}{https://github.com/zsyzzsoft/co-mod-gan}. We use the pre-trained network for image inpainting on FFHQ at resolution 512x512. We downsample the generated images to 256x256 for evaluation.

\section{Qualitative results on image compositon}
\begin{figure*}
\centering
    \subfloat[]{\includegraphics[width=0.18 \linewidth ]{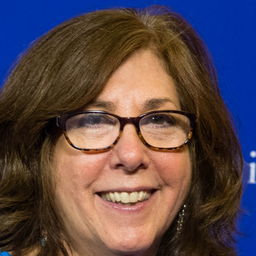}} 
    \subfloat[]{\includegraphics[width=0.18 \linewidth ]{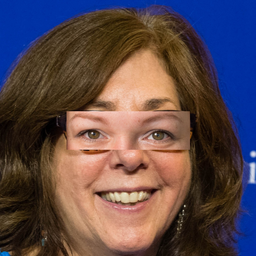}} 
    \subfloat[]{\includegraphics[width=0.18 \linewidth ]{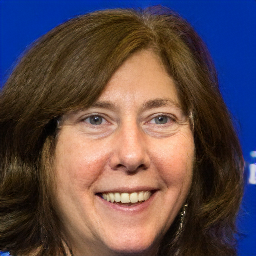}}
    \subfloat[]{\includegraphics[width=0.18 \linewidth ]{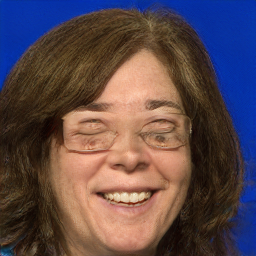}} 
    \subfloat[]{\includegraphics[width=0.18 \linewidth ]{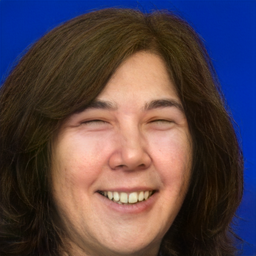}} \\ \vspace{-0.8cm}
     \subfloat[]{\includegraphics[width=0.18 \linewidth ]{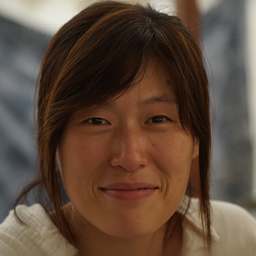}}
      \subfloat[]{\includegraphics[width=0.18 \linewidth ]{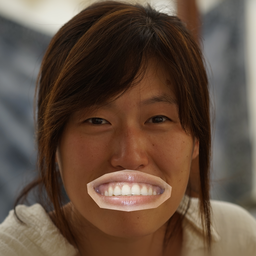}} 
      \subfloat[]{\includegraphics[width=0.18 \linewidth ]{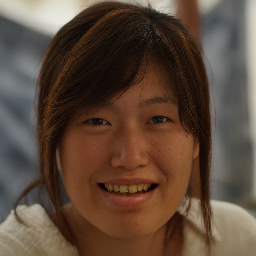}}
      \subfloat[]{\includegraphics[width=0.18 \linewidth ]{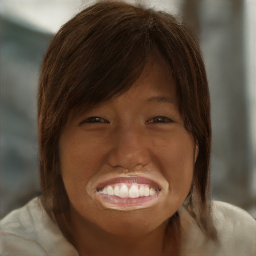}}
       \subfloat[]{\includegraphics[width=0.18 \linewidth ]{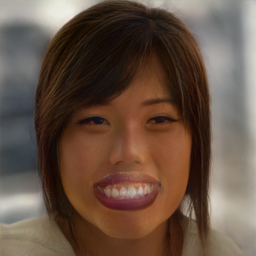}}  \\ \vspace{-0.8cm}
        \subfloat[]{\includegraphics[width=0.18 \linewidth ]{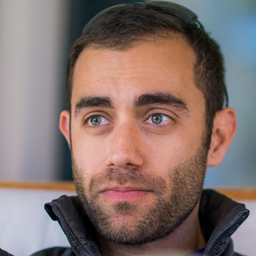}}
      \subfloat[]{\includegraphics[width=0.18 \linewidth ]{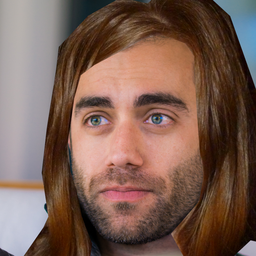}} 
      \subfloat[]{\includegraphics[width=0.18 \linewidth ]{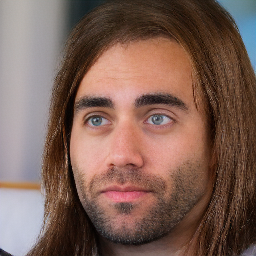}}
      \subfloat[]{\includegraphics[width=0.18 \linewidth ]{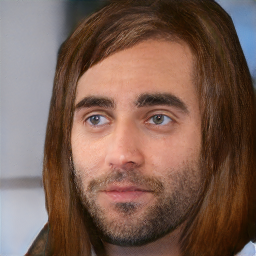}}
       \subfloat[]{\includegraphics[width=0.18 \linewidth ]{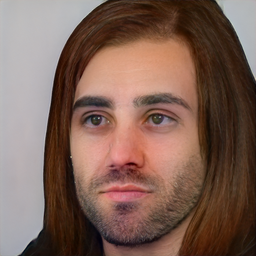}}  \\ \vspace{-0.8cm}
     \subfloat[]{\includegraphics[width=0.18 \linewidth ]{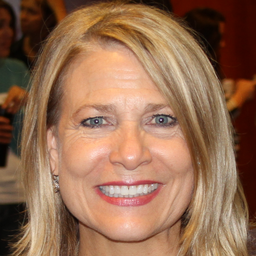}}
     \subfloat[]{\includegraphics[width=0.18 \linewidth ]{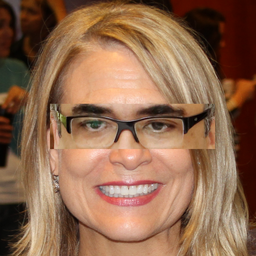}}
     \subfloat[]{\includegraphics[width=0.18 \linewidth ]{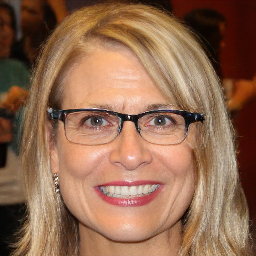}}
      \subfloat[]{\includegraphics[width=0.18 \linewidth ]{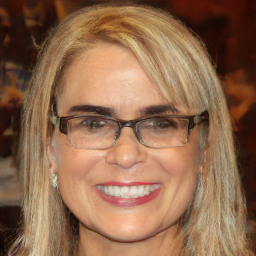}}
     \subfloat[]{\includegraphics[width=0.18 \linewidth ]{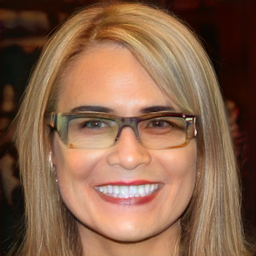}} \\ \vspace{-0.8cm}
    \subfloat[]{\includegraphics[width=0.18 \linewidth ]{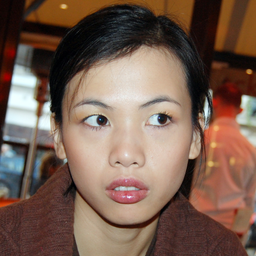}}
    \subfloat[]{\includegraphics[width=0.18 \linewidth ]{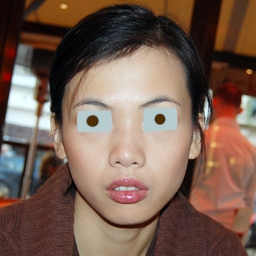}}
    \subfloat[]{\includegraphics[width=0.18 \linewidth ]{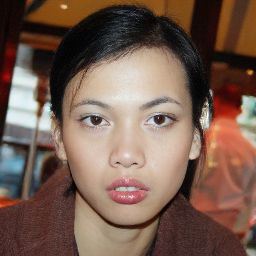}}
     \subfloat[]{\includegraphics[width=0.18 \linewidth ]{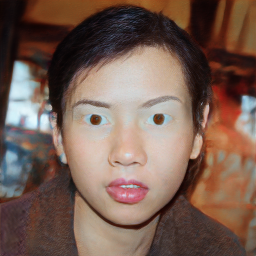}}
    \subfloat[]{\includegraphics[width=0.18 \linewidth ]{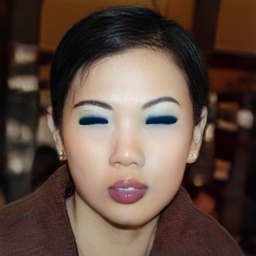}}
    \\ \vspace{-0.8cm}
    \subfloat[]{\includegraphics[width=0.18 \linewidth ]{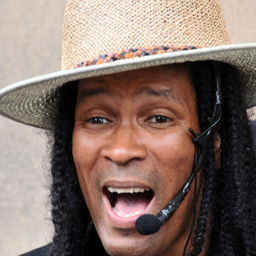}}
    \subfloat[]{\includegraphics[width=0.18 \linewidth ]{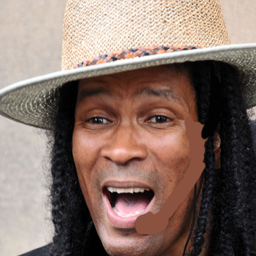}}
    \subfloat[]{\includegraphics[width=0.18 \linewidth ]{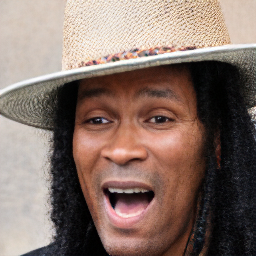}}
     \subfloat[]{\includegraphics[width=0.18 \linewidth ]{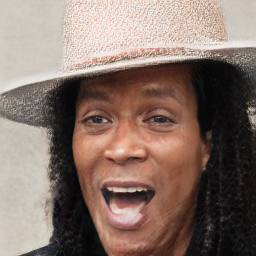}}
    \subfloat[]{\includegraphics[width=0.18 \linewidth ]{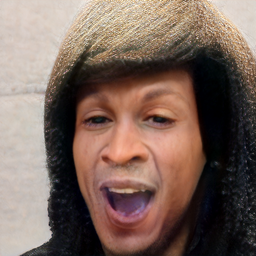}} \\ \vspace{-0.8cm}
    \subfloat[Source]{\includegraphics[width=0.18 \linewidth ]{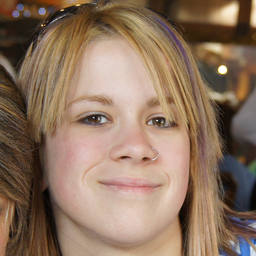}}
    \subfloat[Composite]{\includegraphics[width=0.18 \linewidth ]{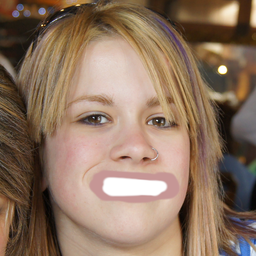}}
    \subfloat[EdiBERT]{\includegraphics[width=0.18 \linewidth ]{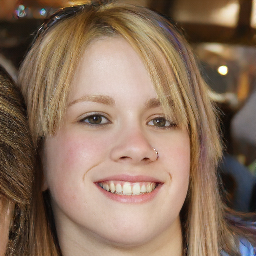}}
    \subfloat[ID-GAN \cite{zhu2020domain}]{\includegraphics[width=0.18 \linewidth ]{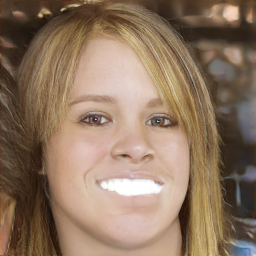}}
    \subfloat[I2SG$\dagger$++ \cite{abdal2020image2stylegan++,karras2020training}]{\includegraphics[width=0.18 \linewidth ]{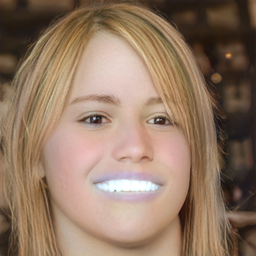}}
     \\
    
    \caption{Image compositing and scribble-based editing on FFHQ.\label{fig:sup_compos_ffhq}}
\end{figure*}

\begin{figure*}
\centering
    \subfloat[]{\includegraphics[width=0.18 \linewidth ]{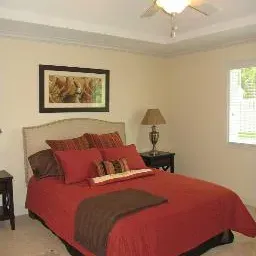}}
    \subfloat[]{\includegraphics[width=0.18 \linewidth ]{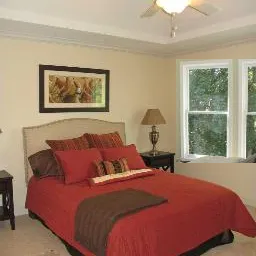}}
    \subfloat[]{\includegraphics[width=0.18 \linewidth ]{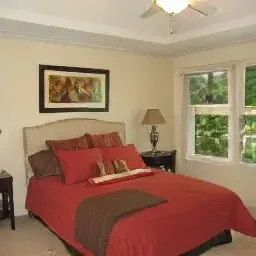}}
     \subfloat[]{\includegraphics[width=0.18 \linewidth ]{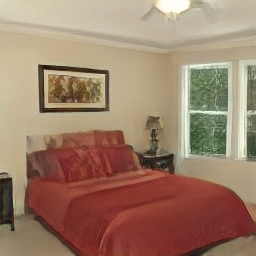}}
    \subfloat[]{\includegraphics[width=0.18 \linewidth ]{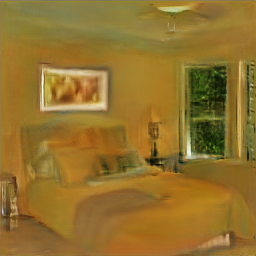}} \\ \vspace{-0.8cm}
    \subfloat[]{\includegraphics[width=0.18 \linewidth ]{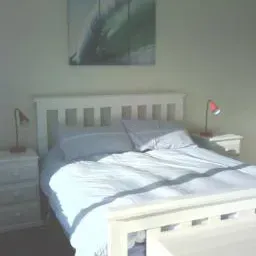}}
    \subfloat[]{\includegraphics[width=0.18 \linewidth ]{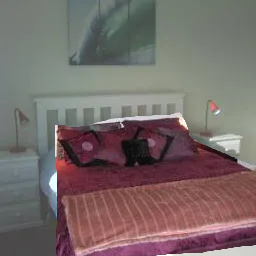}}
    \subfloat[]{\includegraphics[width=0.18 \linewidth ]{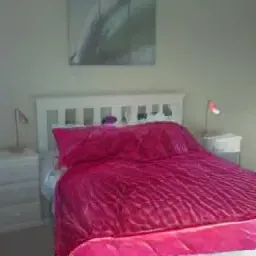}}
    \subfloat[]{\includegraphics[width=0.18 \linewidth ]{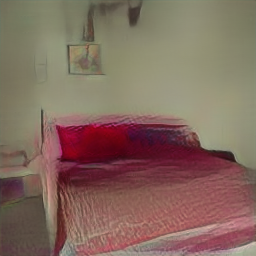}}
    \subfloat[]{\includegraphics[width=0.18 \linewidth ]{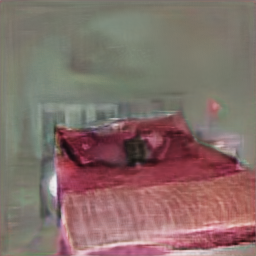}} \\ \vspace{-0.8cm}
    \subfloat[]{\includegraphics[width=0.18 \linewidth ]{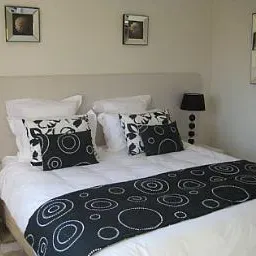}}
    \subfloat[]{\includegraphics[width=0.18 \linewidth ]{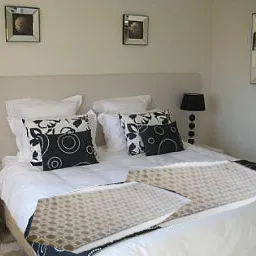}}
    \subfloat[]{\includegraphics[width=0.18 \linewidth ]{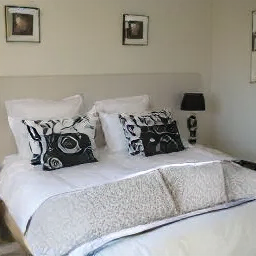}}
     \subfloat[]{\includegraphics[width=0.18 \linewidth ]{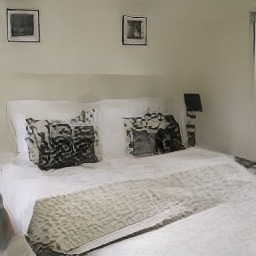}}
    \subfloat[]{\includegraphics[width=0.18 \linewidth ]{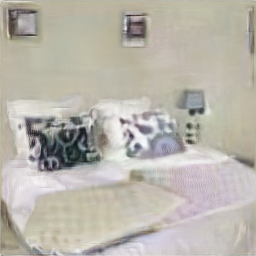}}  \\ \vspace{-0.8cm}
    \subfloat[]{\includegraphics[width=0.18 \linewidth ]{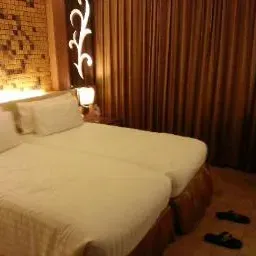}}
    \subfloat[]{\includegraphics[width=0.18 \linewidth ]{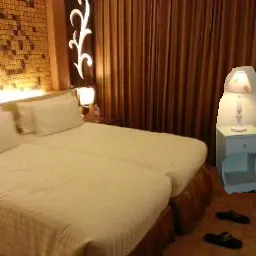}}
     \subfloat[]{\includegraphics[width=0.18 \linewidth ]{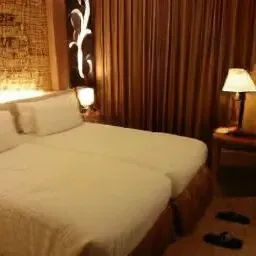}} 
      \subfloat[]{\includegraphics[width=0.18 \linewidth ]{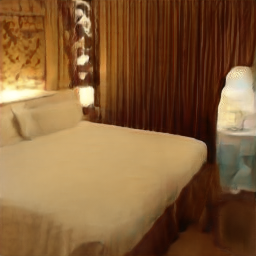}}
    \subfloat[]{\includegraphics[width=0.18 \linewidth ]{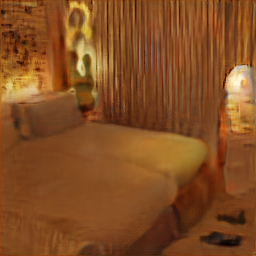}} \\ \vspace{-0.8cm}
    \subfloat[]{\includegraphics[width=0.18 \linewidth ]{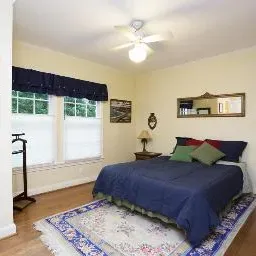}}
    \subfloat[]{\includegraphics[width=0.18 \linewidth ]{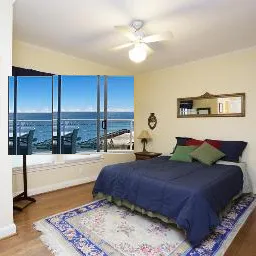}}
    \subfloat[]{\includegraphics[width=0.18 \linewidth ]{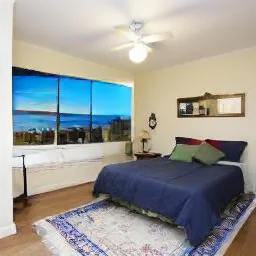}} 
    \subfloat[]{\includegraphics[width=0.18 \linewidth ]{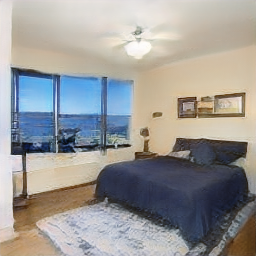}}
    \subfloat[]{\includegraphics[width=0.18 \linewidth ]{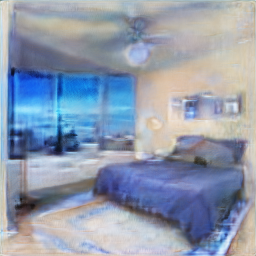}} \\ \vspace{-0.8cm}
    \subfloat[Source]{\includegraphics[width=0.18 \linewidth ]{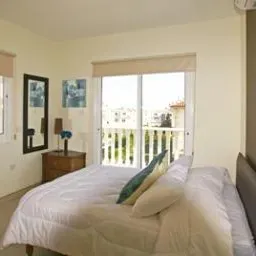}}
    \subfloat[Composite]{\includegraphics[width=0.18 \linewidth ]{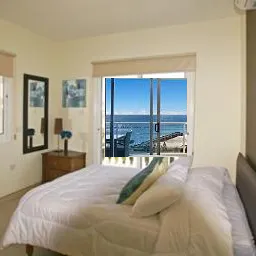}}
    \subfloat[EdiBERT]{\includegraphics[width=0.18 \linewidth ]{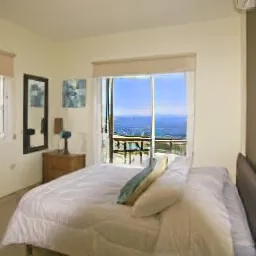}}
    \subfloat[ID-GAN \cite{zhu2020domain}]{\includegraphics[width=0.18 \linewidth ]{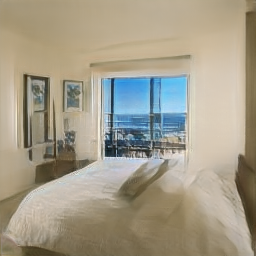}}
    \subfloat[I2SG \cite{abdal2019image2stylegan}]{\includegraphics[width=0.18 \linewidth ]{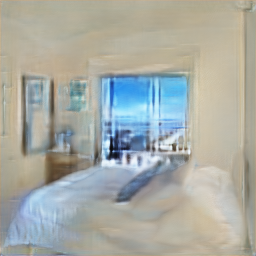}} \\
    \caption{Image compositing on LSUN Bedroom. \label{fig:sup_compos_lsun}}
\end{figure*}

We present more examples of image compositions, with image compositing and scribble-based editing on FFHQ in Figure \ref{fig:sup_compos_ffhq}, and image compositing on LSUN Bedroom in Figure \ref{fig:sup_compos_lsun}.

\textbf{Preservation of non-masked parts.} Thanks to its VQGAN auto-encoder, EdiBERT generally better conserves areas outside the mask than GANs inversion methods. This is particularly visible for images with complex backgrounds on FFHQ (Figure \ref{fig:sup_compos_ffhq}, 5th and last rows).

\textbf{Insertion of edited parts.} Since EdiBERT is a probabilistic model and the tokens inside the modified area are resampled, the inserted object can be modified and mapped to a more likely object given the context. It thus generates more realistic images, but can alter the fidelity to the inserted object. For example, on row 1 of Figure \ref{fig:sup_compos_lsun}, the green becomes lighter and the perspective of the inserted window is improved. Although it can be a downside for image compositing, note that this property is interesting for scribble-based editing, where the scribbles have to be largely transformed to get a realistic image. Contrarily, GANs inversion methods tend to conserve the inserted object too much, even if it results in a highly unrealistic generated image. We can observe this phenomenon on last row of Figure \ref{fig:sup_compos_ffhq}.

\captionsetup[subfigure]{labelformat=empty}

\captionsetup[subfigure]{labelformat=empty}

\clearpage

\section{Survey on FFHQ image compositing}

The survey was presented as a Google Form with 40 questions. For each question, the user was shown 6 images: Source, Composite, EdiBERT, ID-GAN \cite{zhu2020domain},  I2SG$\dagger$++ \cite{abdal2019image2stylegan,karras2020training}, LC \cite{chai2021using}.  The different generated images were referred as Algorithm 1, ..., Algorithm 4. The user was asked to vote for its preferred generated image, by taking into account realism and fidelity criterions. The user had no time limit for the poll. 30 users answered our poll. We provide the detailed answers for each image in Table \ref{table:detailUserStudy}.

\begin{table}[b]
\centering
\begin{tabular}{c|c|c|c|c|}
& EdiBERT & ID-GAN \cite{zhu2020domain} & LC \cite{chai2021using} & I2SG$\dagger$++ \cite{abdal2019image2stylegan,karras2020training} \\ \hline
& 17 & 5 & 7 & 1 \\
& 15 & 1 & 8 & 6 \\
& 22 & 4 & 1 & 3 \\ 
& 19 & 4 & 5 & 2 \\ 
& 22 & 0 & 4 & 4 \\
& 6 & 7 & 8 & 9 \\
& 21 & 1 & 5 & 3 \\
& 23 & 1 & 4 & 2 \\
& 20 & 5 & 5 & 0 \\
& 11 & 13 & 6 & 0 \\
& 27 & 0 & 3 & 0 \\
& 12 & 3 & 3 & 12 \\
& 16 & 4 & 6 & 4 \\ 
& 25 & 2 & 1 & 2 \\
& 18 & 8 & 1 & 3 \\
& 8 & 13 & 9 & 0 \\
& 26 & 0 & 4 & 0 \\
& 7 & 0 & 21 & 2 \\
& 14 & 9 & 1 & 6 \\
& 27 & 0 & 1 & 2 \\
& 11 & 19 & 0 & 0 \\
& 14 & 9 & 4 & 3 \\
& 16 & 14 & 0 & 0 \\
& 21 & 1 & 3 & 5 \\
& 8 & 2 & 18 & 2 \\
& 19 & 3 & 3 & 5 \\
& 22 & 7 & 0 & 1 \\
& 23 & 2 & 1 & 4 \\
& 18 & 0 & 2 & 10 \\
& 27 & 2 & 1 & 0 \\
& 22 & 2 & 1 & 5 \\
& 24 & 0 & 5 & 1 \\
& 3 & 25 & 2 & 0 \\
& 28 & 0 & 2 & 0 \\
& 24 & 0 & 6 & 0 \\
& 27 & 0 & 3 & 0 \\
& 27 & 1 & 2 & 0 \\
& 22 & 7 & 1 & 0 \\
& 9 & 15 & 6 & 0 \\
& 14 & 0 & 14 & 2 \\ \hline
Total & 735 (61.25\%) & 189  (15.75\%) & 177 (14.75\%) & 99 (0.0825\%)  \\ 
\end{tabular}
\caption{Detailed results of the user study. Each line corresponds to an image, with the associated number of votes per method.\label{table:detailUserStudy}}
\end{table}

\end{document}